\documentclass[final]{cvpr}

\usepackage{times}
\usepackage{epsfig}
\usepackage{graphicx, caption}
\usepackage{amsmath}
\usepackage{amssymb}

\newcommand{\fig}[1]{Figure~\ref{fig:#1}}
\newcommand{\sect}[1]{Section~\ref{sect:#1}}
\newcommand{\tab}[1]{Table~\ref{tab:#1}}

\newcommand{\eq}[1]{(\ref{eq:#1})}

\usepackage[pagebackref=true,breaklinks=true,colorlinks,bookmarks=false]{hyperref}
\usepackage{array}
\newcommand\tablescale{0.86}

\usepackage[pagebackref=true,breaklinks=true,colorlinks,bookmarks=false]{hyperref}

\def\cvprPaperID{5478} 
\def\confYear{CVPR 2021}

\begin{document}
\title{Navigating the GAN Parameter Space for Semantic Image Editing}

\author{Anton Cherepkov\\
Yandex, Russia\\
Moscow Institute of Physics and Technology\\
{\tt\small cherepkov.ayu@phystech.edu}
\and
Andrey Voynov\\
Yandex, Russia\\
{\tt\small an.voynov@yandex.ru}
\and
Artem Babenko\\
Yandex, Russia\\
HSE University\\
{\tt\small artem.babenko@phystech.edu}
}

\twocolumn[{
\renewcommand\twocolumn[1][]{#1}
\maketitle
\begin{center}
  \vspace{-14pt}
  \includegraphics[width=0.999\textwidth]{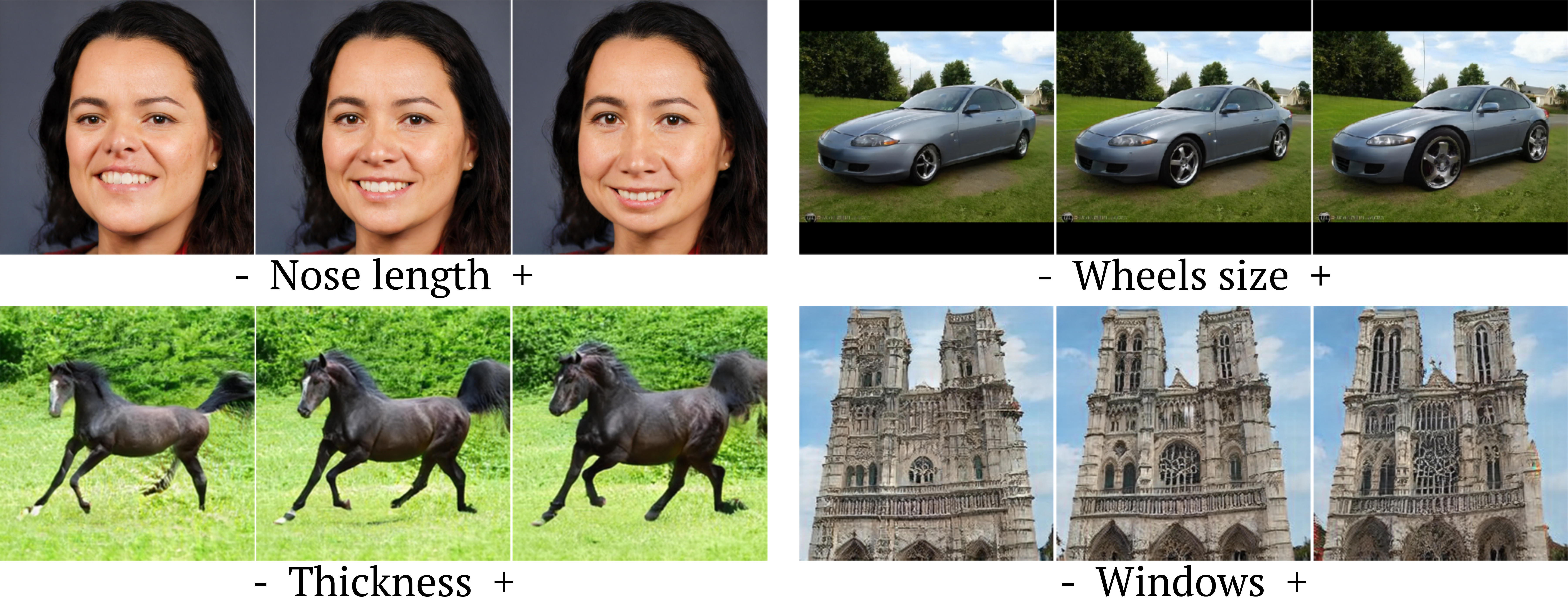}
  \vspace{-12pt}
  \label{fig:teaser}
\end{center}
}]

\let\thefootnote\relax\footnotetext{\hspace{-14pt}Accepted to CVPR 2021\\Contact author: Andrey Voynov, {\tt an.voynov@yandex.ru}}

\begin{abstract}
Generative Adversarial Networks (GANs) are currently an indispensable tool for visual editing, being a standard component of image-to-image translation and image restoration pipelines. Furthermore, GANs are especially advantageous for controllable generation since
their latent spaces contain a wide range of interpretable directions, well suited for semantic editing operations. By gradually changing latent codes along these directions, one can produce impressive visual effects, unattainable without GANs.

In this paper, we significantly expand the range of visual effects achievable with the state-of-the-art models, like StyleGAN2. In contrast to existing works, which mostly operate by latent codes, we \textbf{ discover interpretable directions in the space of the generator parameters}. By several simple methods, we explore this space and demonstrate that it also contains a plethora of interpretable directions, which are an excellent source of non-trivial semantic manipulations. The discovered manipulations cannot be achieved by transforming the latent codes and can be used to edit both synthetic and real images. We release our code and models\footnote{\url{https://github.com/yandex-research/navigan}} and hope they will serve as a handy tool for further efforts on GAN-based image editing.
\vspace{-3mm}
\end{abstract}

\section{Introduction}
\label{sect:intro}

Generative Adversarial Networks (GANs) \cite{goodfellow2014generative} have revolutionized image processing research, significantly pushing the boundaries of machine learning for image enhancement and visual editing. Different research lines currently exploit GANs in several principled ways, e.g. using them as an implicit learnable objective \cite{li2016precomputed, ledig2017photo, isola2017image, choi2018stargan, lee2020drit++, zhu2017unpaired, zhu2017toward}, employing them as high-quality image priors \cite{zhu2016generative, bau2019semantic, gu2020image, menon2020pulse, pan2020exploiting}, manipulating their internal representations for visual editing purposes \cite{bau2018gan, collins2020editing}. Furthermore, the GAN latent spaces often encode human-interpretable concepts \cite{radford2015unsupervised, shen2020interpreting, goetschalckx2019ganalyze, jahanian2019steerability, plumerault2019Controlling, voynov2020unsupervised, harkonen2020ganspace, peebles2020hessian, wang2021a}, which makes GANs the dominant paradigm for controllable generation.

Since the seminal paper \cite{radford2015unsupervised},  which has demonstrated the semantic arithmetic of latent vectors in GANs, plenty of methods to discover interpretable directions in the GAN latent spaces have been developed \cite{radford2015unsupervised, shen2020interpreting, goetschalckx2019ganalyze, jahanian2019steerability, plumerault2019Controlling, voynov2020unsupervised, harkonen2020ganspace, peebles2020hessian, wang2021a}.
These methods successfully identify such directions across different GAN models and hold great potential for effective image editing. These days, many impressive visual effects can be achieved by simply moving the image latent codes along these directions.

Our paper demonstrates that a large number of exciting non-trivial visual effects can be produced by gradually \textit{modifying the GAN parameters rather than the latent codes}. In more detail, we show that the GAN parameter space also contains a plethora of directions, corresponding to interpretable image manipulations. Moreover, we describe simple domain-agnostic procedures that discover such directions in an unsupervised fashion. By extensive experiments, we confirm that the discovered visual effects are substantially new and cannot be achieved by the latent code manipulations. Overall, our findings significantly expand the arsenal of GAN-based image editing techniques.

To sum up, our contributions are the following:

\begin{itemize}
    \item We propose to use the interpretable directions in the space of the generator parameters for semantic editing. Our approach differs from existing works, which operate by the latent codes or the intermediate GAN activations. Our findings demonstrate that remarkable visual effects can be achieved by slightly changing the GAN parameters.
    
    \item We develop the methods to discover such directions. The proposed methods are both effective and fast and can work on a single GPU.
    
    \item We confirm that the discovered directions are qualitatively new and correspond to semantic manipulations, which existing methods cannot produce.
\end{itemize}

\section{Related work}
\label{sect:related}

This section describes the typical ways to exploit GANs for visual editing purposes and positions our work with respect to existing literature.

\textbf{GAN discriminators as learnable training objectives.} In most image generation tasks, it is challenging to explicitly define an objective term that would enforce the produced images' realism. Therefore, many existing approaches employ GANs as implicit learnable objectives, making the output images indistinguishable from the real ones. This approach has become de-facto standard for a wide range of image processing tasks, including super-resolution \cite{ledig2017photo}, texture transfer \cite{li2016precomputed}, image-to-image translation \cite{isola2017image, choi2018stargan, lee2020drit++, zhu2017unpaired, zhu2017toward}.

\textbf{GANs as high-quality image priors.} As the state-of-the-art GAN models are high-quality approximations of the real image manifold, several recent methods use these models as ``hard'' priors. In this case, the method's outputs are produced by a large-scale pretrained GAN, thus, the image processing task is reduced to the optimization problem in the GAN latent space. This paradigm has been successfully used for super-resolution \cite{menon2020pulse}, visual editing \cite{zhu2016generative, bau2019semantic}, image restoration tasks \cite{gu2020image, pan2020exploiting}, e.g. colorization and inpainting.

\textbf{Latent manipulations in GANs.} The prior literature has empirically shown that the GAN latent spaces are endowed with human-interpretable vector space arithmetic \cite{radford2015unsupervised, shen2020interpreting, goetschalckx2019ganalyze, jahanian2019steerability, voynov2020unsupervised}. E.g., in GANs trained on face images, their latent spaces possess linear directions corresponding to adding smiles, glasses, gender swap \cite{radford2015unsupervised, shen2020interpreting}. Since such interpretable directions provide a straightforward route to powerful image editing, their discovery currently receives much research attention. A line of recent works \cite{goetschalckx2019ganalyze, shen2020interpreting} employs explicit human-provided supervision to identify interpretable directions in the latent space. For instance, \cite{shen2020interpreting} use the classifiers pretrained on the CelebA dataset \cite{liu2015deep} to predict certain face attributes. These classifiers are then used to produce pseudo-labels for the generated images and their latent codes. Based on these pseudo-labels, the separating hyperplane is constructed in the latent space, and a normal to this hyperplane becomes a direction, controlling the corresponding attribute. Another work \cite{goetschalckx2019ganalyze} solves the optimization problem in the latent space, maximizing the score of the pretrained model, which predicts image memorability. The result of this optimization is the direction ``responsible'' for the increase of memorability. Two self-supervised works \cite{jahanian2019steerability, plumerault2019Controlling} seek the vectors in the latent space that correspond to simple image augmentations such as zooming or translation. Finally, a bunch of recent methods \cite{voynov2020unsupervised, harkonen2020ganspace, peebles2020hessian} identify interpretable directions without any form of supervision. \cite{voynov2020unsupervised} learns a set of directions that can be easily distinguished by a separate classification model based on two samples, produced from the original latent codes and their versions shifted along the particular direction. \cite{peebles2020hessian} learns the directions by minimizing the sum of squared off-diagonal terms of the generator's Hessian matrix. Another approach, \cite{harkonen2020ganspace}, 
demonstrates that interpretable directions often correspond to the principal components of the activations from the first layer of the generator network. A very recent work \cite{wang2021a} aims to unify all unsupervised approaches above. \cite{wang2021a} claims that these approaches can be treated as special cases of computing the spectrum of the Hessian for the LPIPS model \cite{zhang2018unreasonable} with respect to latent coordinates. Intuitively, \cite{wang2021a} assumes that the interpretable directions correspond to the largest perceptual changes, quantified by the LPIPS model. Our work is partially inspired by the ideas from the latent space exploration, however, we explore the space of the generator parameters rather than the latent space.

\textbf{Manipulating GAN activations.} It was shown \cite{bau2018gan} that the intermediate activations in the GAN generator often correspond to various semantic concepts. \cite{bau2018gan} exploits this by controlling the presence or absence of objects at given positions, guided by supervision from a pretrained semantic segmentation model. Another unsupervised approach \cite{collins2020editing} identifies semantically meaningful activations via simple k-means clustering and makes local edits for the generated images. In contrast to these works, we operate on the generator weights rather than the activations tensors.

\textbf{Rewriting a deep generative model.} The closest work to ours is \cite{bau2020rewriting}, which shows that a user-specified editing operation can be achieved by changing the generator weights. In a nutshell, \cite{bau2020rewriting} solves the optimization problem that seeks the shift in the generator's parameter space, such that the modified generator would produce the images after a needed editing operation. In contrast, our approach does not require specifying operations a priori and discovers a lot of unexpected non-trivial visual effects, some of which are impossible to formulate explicitly.

\section{Methods}
\label{sect:method}

\subsection{Preliminaries}

Assume that we have a pretrained GAN generator $G_{\theta}$, which maps the samples $z \sim \mathcal{N}(0,\mathbb{I})$ from the latent space $\mathbb{R}^{d}$ into the image space,  $\mathbb{R}^{W\times H \times 3}$ $G: z \rightarrow I$, and ${\theta} \in \Theta$ denotes the set of the generator's parameters. Our goal, then, is to learn a set of vectors $\xi_1,\dots,\xi_K \in \Theta$, such that changing the generator parameters along these vectors effectively performs continious semantic editing operations. More formally, the mappings $G_{\theta}(z) \rightarrow G_{\theta + t\xi_k}(z), k{=}1..K$ have to correspond to interpretable visual effects, consistent for all latent codes $z \sim \mathcal{N}(0,\mathbb{I})$. Here $t \in \left[-T,T \right]$ is a shift magnitude, which controls the degree of the corresponding visual effect.

Note that the problem statement above resembles the established problem of learning the interpretable latent controls addressed in \cite{shen2020interpreting, jahanian2019steerability, voynov2020unsupervised, harkonen2020ganspace, wang2021a}. However, the existing works perform the editing operations by shifting the latent codes with a fixed generator. In contrast, we operate in a much higher dimensional space of the generator's parameters $\Theta$ and do not change the latent codes when editing. In the next subsections, we describe the methods that discover the interpretable directions in $\Theta$.

\subsection{Optimization-based approach}
\label{sect:optimization}

The first approach is inspired by the recent unsupervised technique \cite{voynov2020unsupervised}, which discovers the interpretable latent directions. Intuitively,
\cite{voynov2020unsupervised} assumes that interpretable directions are the ones that are easy to distinguish from each other, observing only the results of the corresponding image manipulations. Here, we build on this intuition to explore the parameter space $\Theta$. While many parts of our protocol have been used in \cite{voynov2020unsupervised}, we still present them for self-containedness.

\begin{figure}[h!]
    \centering
    \includegraphics[width=\columnwidth]{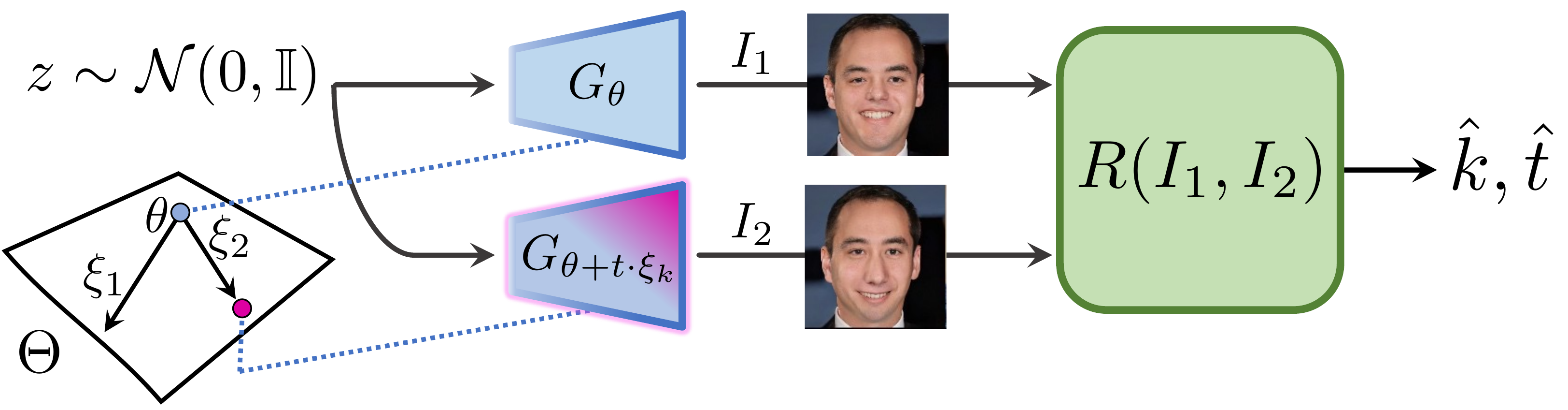}
    \caption{Our learning protocol that discover the interpretable shifts in the space of the generator's parameters $\Theta$. A training sample consists of two images, produced by the generators with original and shifted parameters. The images are given to a reconstructor $R$ that predicts a direction index $k$ and a shift magnitude~$t$.}
    \vspace{-2mm}
    \label{fig:rectification}
\end{figure}

\fig{rectification} demonstrates our training scheme. It includes two learnable modules:

\begin{enumerate}
    \item Direction matrix $\Xi{=}\left[\xi_1,\dots,\xi_K\right] \in \mathbb{R}^{dim(\Theta)\times K}$.
    \item Reconstructor $R$, which recieves the results of editing operations represented by pairs of the form $\{G_{\theta}(z); G_{\theta + t\xi_k}(z)\}$, and predicts both $k$ and $t$ from its input. More formally, $R$ is a function $(I_1,\ I_2)\longrightarrow(\{1,\dots,K\},\ \mathbb{R})$ parameterized by a deep convolutional network.
\end{enumerate}

The learning is performed by minimizing the expected reconstructor's prediction error:

\vspace{-2mm}
\begin{equation}
\begin{split}
\underset{\left[\xi_1,\dots,\xi_K\right],R}{\text{\large{min}}}\hspace{2mm}\underset{\substack{z\sim\mathcal{N}(0,\mathbb{I}) \\ k\sim\mathcal{U}\{1,K\} \\ t\sim\mathcal{U}\left[-T,T\right]}}{\mathbb{E}}\left[L_{cl}(k,\widehat{k}) + \lambda L_r (t, \widehat{t})\right]
\end{split}
\label{eq:L_total}
\end{equation}
\vspace{-2mm}

where $\widehat{k}$ and $\widehat{t}$ denote the reconstructor's output:

\vspace{-2mm}
\begin{equation}
    (\widehat{k}; \widehat{t})= R\left(G_{\theta}(z); G_{\theta + t\xi_k}(z)\right)
\end{equation}
\vspace{-2mm}

For the classification objective term $L_{cl}(\cdot,\cdot)$ we use cross-entropy, and for the regression term $L_{r}(\cdot,\cdot)$ mean absolute error is used. Since all the components of our scheme are differentiable, it can be optimized jointly by stochastic gradient descent.

\begin{figure}[h]
    \centering
    \includegraphics[width=0.7\columnwidth]{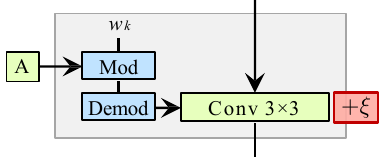}
    \caption{The additive shift $\xi$ is added to the convolutional kernel weight in the StyleGAN2 demodulation block.}
    \label{fig:conv_block}
    \vspace{-3mm}
\end{figure}

\textbf{Reducing the dimensionality of the optimization space.} State-of-the-art generators (e.g., StyleGAN2) typically have millions of parameters $\theta$, making it infeasible to learn the full $\Xi \in \mathbb{R}^{dim(\Theta)\times K}$ matrix. To simplify the optimization, in experiments, we minimize \eq{L_total} considering only the shifts $\xi_1,\dots,\xi_K$ applied
to a particular generator's layer, with all other parameters being fixed. Namely, we add a shift to the convolutional kernel of one StyleGAN2 block (see \fig{conv_block}). This design choice is motivated by the evidence from \cite{bau2018gan} that different generator layers typically affect the different aspects of the image.

However, only considering a single layer does not fully solve the problem of high dimensionality. To reduce the optimization space even further, we perform the following. Inspired by the recent work on few-shot GAN adaptation \cite{robb2020few}, we
compute the SVD decomposition of the chosen convolutional layer, flattened to a 2D matrix, as in \cite{robb2020few}. Then we optimize over the shifts $\xi_1,\dots,\xi_K$ applied only to the singular values of the diagonal matrix from SVD. Such parametrization makes the optimization problem feasible but is also expressive enough to represent various visual effects, as shown experimentally. While we tried some alternative parameterizations, they typically resulted in worse visual performance.

\vspace{1mm}
\textbf{Practical details.}
\vspace{-1mm}
\begin{enumerate}
    \item In all experiments we use the Resnet-18 architecture \cite{he2016deep} for the reconstructor, with two heads, predicting $t$ and $k$ respectively. We downscale the input images to the resolution $256 \times 256$ to decrease memory consumption.
    \vspace{-1mm}
    \item Assuming that SVD of a flattened convolutional kernel equals $U \cdot \mathrm{diag}(\sigma_1, \dots, \sigma_n) \cdot V$, we apply an additive shift $\xi = (\xi^{(1)}, \dots. \xi^{(n)})$ to the singular values, mapping $\mathrm{diag}(\sigma_1, \dots, \sigma_n)$ to $\mathrm{diag}(\sigma_1 + \xi^{(1)}, \dots, \sigma_n + \xi^{(n)})$. We also normalize the shift $\xi$ to a unit length to avoid parameter explosion.
    \vspace{-1mm}
    \item In all experiments we use $K{=}64$,  $\lambda{=}2.5 \cdot 10^{-3}$ and $T{=}3500$. For optimization, we use Adam with a constant learning rate $0.0001$ and perform $10^5$ learning iterations with a batch size $32$.
\end{enumerate}

\subsection{Spectrum-based approach}
\label{sect:spectrum}

Our second approach originates from an alternative premise presented in \cite{wang2021a}, which claims that the interpretable directions in the latent space should correspond to as large perceptual changes of the generated images as possible. \cite{wang2021a} formalizes this intuition by the following. They calculate the latent directions that are the eigenvectors of the Hessian of the LPIPS model \cite{zhang2018unreasonable} computed with respect to the latent codes $z$. \cite{wang2021a} also proposes an efficient way to compute top-k eigenvectors, corresponding to the largest eigenvalues, avoiding the explicit computation of the whole Hessian, which is impractical.

Our second approach exploits the same intuition as in \cite{wang2021a} but operates with the Hessian of LPIPS computed with respect to the generator's parameters. For completeness, we briefly describe the main steps to compute the Hessian's top-eigenvectors, though they are almost the same as in \cite{wang2021a}. Let us denote by $d(\cdot,\cdot)$ the LPIPS model, which is shown to capture the perceptual distance between two images \cite{zhang2018perceptual}. Let $\theta$ be the weights of the pretrained GAN generator . Then, we consider the quantity $\mathbb{E}_z d^2(G_{\theta}(z), G_{\theta + \alpha}(z))$, which is the expected perceptual change induced by shifting the generator parameters by $\alpha$. Assuming the LPIPS smoothness, we can write
\vspace{-1mm}
\begin{multline}
    \mathbb{E}_z d^2(G_{\theta}(z), G_{\theta + \delta\alpha}(z)) = \\
    \mathbb{E}_z  d^2(G_{\theta}(z), G_{\theta}(z)) + \frac{\partial d^2(G_{\theta}(z), G_{\theta + \alpha}(z))}{\partial \alpha}|_{\alpha = 0} \cdot \delta\alpha + \\
    \delta\alpha^T \cdot \frac{\partial^2 d^2(G_\theta(z), G_{\theta + \alpha}(z))}{\partial \alpha^2} |_{\alpha = 0} \cdot \delta\alpha + \bar{o}(\|\delta\alpha\|^2_2)
    \label{eq:hessian_inference}
\end{multline}
The first two terms from the right side of \eq{hessian_inference} are equal to zero since $d^2$ achieves its global minimum at $\alpha = 0$. Thus, we focus on the eighenvectors of the Hessian
$$
H = \frac{\partial^2 \mathbb{E}_zd^2(G_\theta(z), G_{\theta + \alpha}(z))}{\partial \alpha^2} |_{\alpha = 0}
$$
For efficient computation, define the gradient function $g(a) = \frac{\partial\mathbb{E}_z d^2(G_{\theta}(z), G_{\theta + \alpha}(z))}{\partial \alpha}|_{\alpha = a}$. Following \cite{wang2021a}, we sample $v \sim \mathcal{N}(0,\mathcal{I})$ and iteratively update it by the rule:
\begin{equation}
v \mapsto \frac{g(\varepsilon v) - g(-\varepsilon v)}{2 \varepsilon \|v\|}
\label{eq:HVD_fast}
\end{equation}
where $\varepsilon$ is a small constant set to $0.1$. This process converges to the leading eigenvector of the hessian $H$ \cite{lehoucq1998arpack}. Once the top $k - 1$ eigenvectors are found, we repeat this procedure, restricted to their orthogonal complementary, to obtain the $k$-th eigenvector. Namely, on every step of \eq{HVD_fast} we project $v$ into the orthogonal complement of the already found top-$(k - 1)$ eigenvectors.

In the experiments, we compute the top-64 eigenvectors. We approximate the expectation of the gradient function $g$ with a minibatch of $512$ randomly sampled $z \sim \mathcal{N}(0, \mathbb{I})$. We always perform ten iterations of \eq{HVD_fast} since it is sufficient for convergence. As in the optimization-based approach, we explore the parameter subspaces corresponding to different layers separately and compute the eigenvectors only with respect to parameters from a particular layer.

\subsection{Hybrid scheme}
\label{sect:hybrid}

We also propose a simple hybrid scheme that combines both the optimization-based and the spectrum-based approaches in a single procedure. 
First, we compute the top-k eigenvectors of the LPIPS Hessian $v_1,\dots,v_k$, as described in \sect{spectrum}. Then we solve the optimization problem \eq{L_total} considering only the shifts $\xi$ applied to $v_1,\dots,v_k$. Informally, this hybrid scheme is equivalent to the optimization-based approach that operates in the parameter subspace that captures the maximal perceptual differences in the generated images. For the hybrid scheme, we use the same hyperparameter values as in \sect{optimization} except for $T$, which was set to $80$ and batch size $16$ in purpose to be able to run on a single Tesla V100. This change is due to the Hessian directions produce more noticeable visual effects compared to SVD.

\subsection{Inspecting directions.}

The methods from \sect{optimization}, \sect{spectrum}, \sect{hybrid} all produce a set of directions $\xi_1,\dots,\xi_K$. These $K$ directions are then inspected manually by observing the image sequences $\{G_{\theta+t\xi_k}(z) | t \in \left[-T,T\right]\}$, for several latent codes $z$. If a particular sequence maintains the image realism and corresponds to interpretable manipulation, it then can be used in editing applications.
Since $K$ is typically small (e.g., $64$), this procedure takes only several minutes for a single person.
\section{Experiments}
\label{sect:experiments}

In this section, we evaluate the proposed techniques on the state-of-the-art StyleGAN2 generators \cite{Karras2019stylegan2} pretrained on the FFHQ \cite{karras2019style}, LSUN-Cars \cite{yu2015lsun}, LSUN-Horse \cite{yu2015lsun} and LSUN-Church \cite{yu2015lsun} datasets. Our goal is to demonstrate that the StyleGAN2 parameter space contains the linear directions that can be exploited for high-fidelity image editing.

\subsection{Non-trivial visual effects}

First, we present several visual effects induced by the discovered directions. \fig{examples} and \fig{examples2} present three examples for each dataset. More examples are provided in supplementary and at the repository\footnote{\url{https://github.com/yandex-research/navigan}}.

\begin{figure}
    \centering
    \includegraphics[width=\columnwidth]{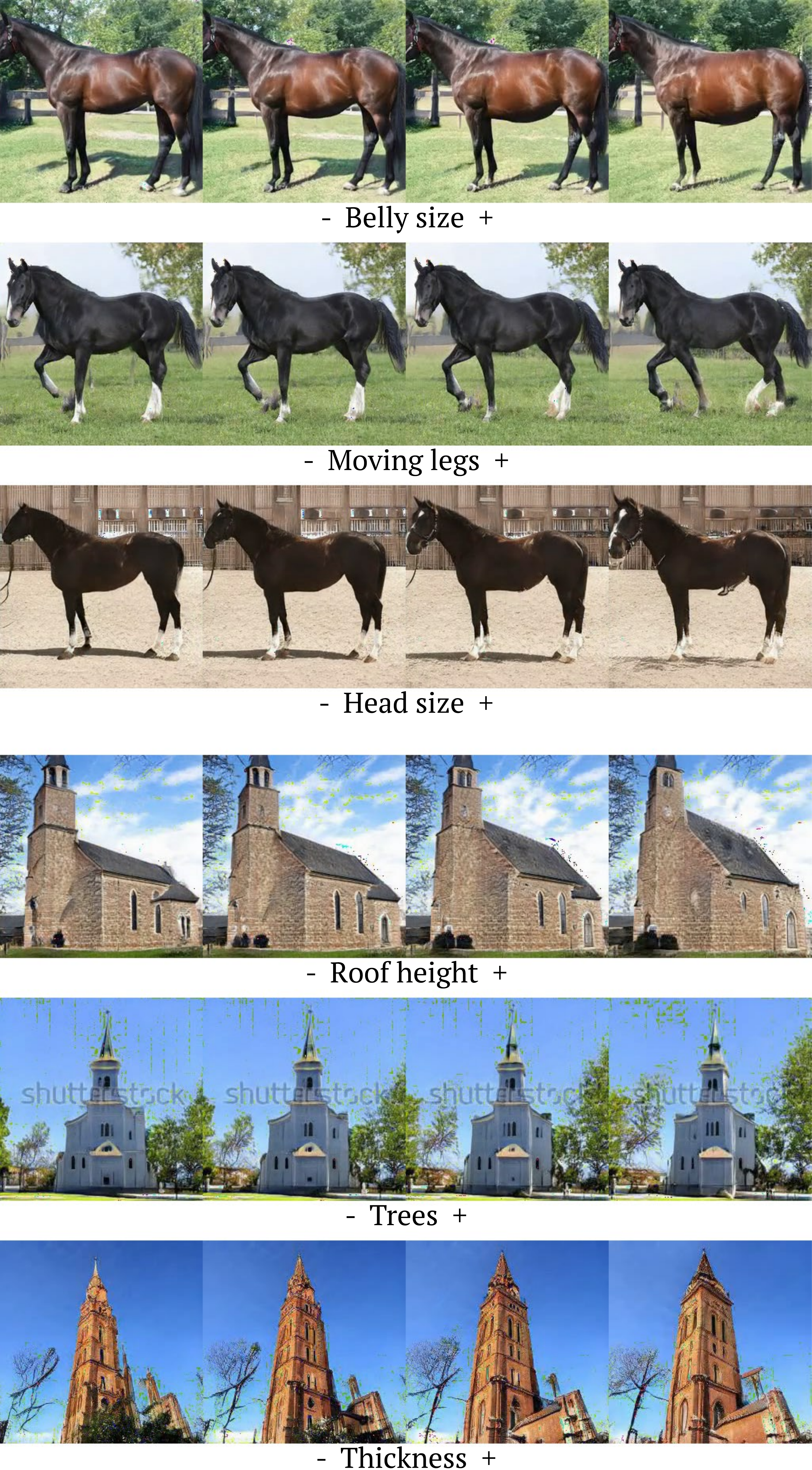}
    \caption{Visual effects achieved by navigating the StyleGAN2 parameter space. Rows $1$-$3$ correspond to  the LSUN-Horse dataset and rows $4$-$6$ correspond to  the LSUN-Church dataset.}
    \vspace{-6mm}
    \label{fig:examples}
\end{figure}

\begin{figure}
    \centering
    \includegraphics[width=\columnwidth]{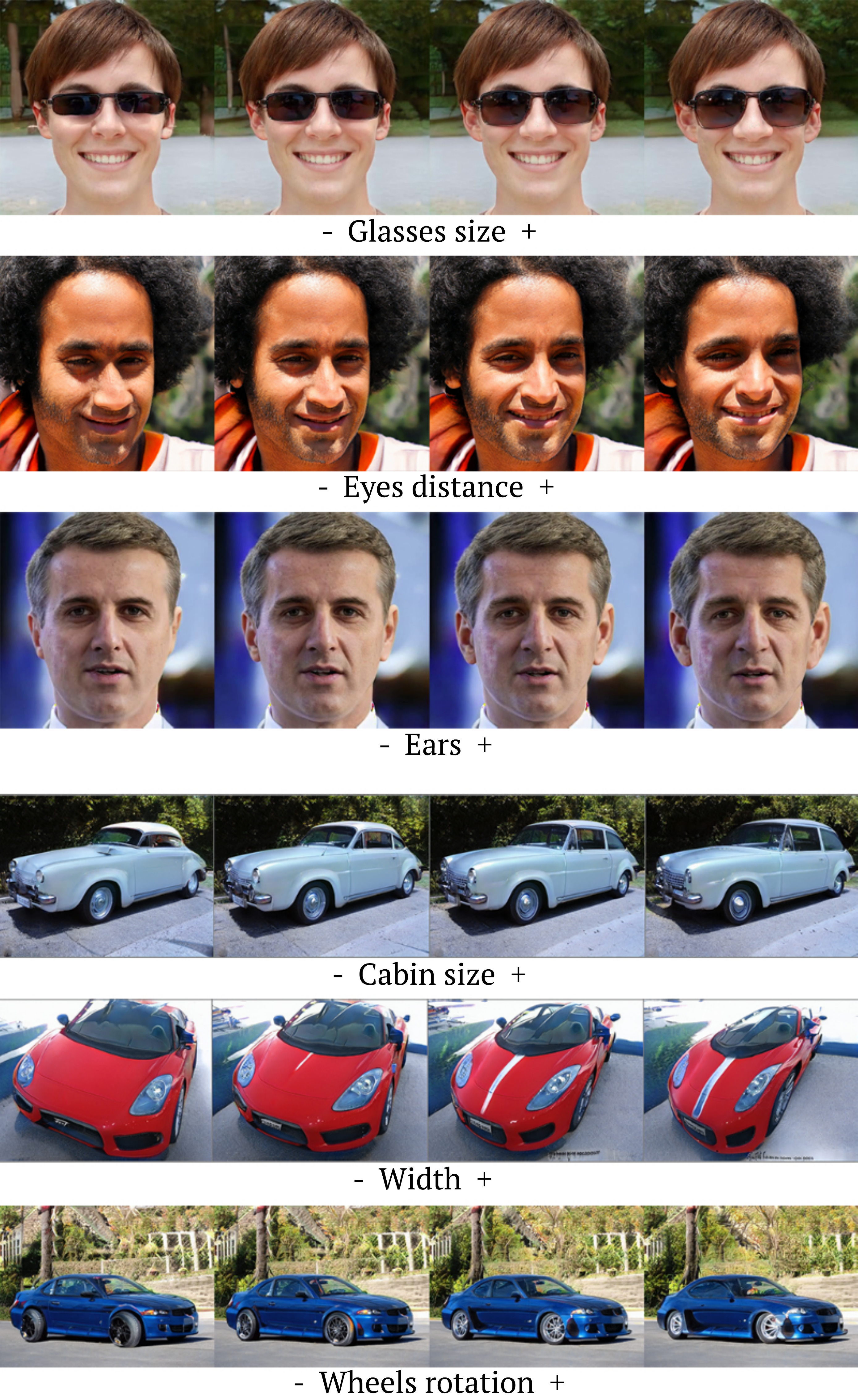}
    \caption{Visual effects achieved by navigating the StyleGAN2 parameter space. Rows $1$-$3$ correspond to  the FFHQ dataset and rows $4$-$6$ correspond to the LSUN-Cars dataset.}
    \vspace{-5mm}
    \label{fig:examples2}
\end{figure}

The typical manipulations affect scales and aspect ratio of certain object parts, like ``Belly size'' and ``Nose length'', or mutual arrangement of these parts, e.g., ``Moving legs'' and ``Wheels rotation''.
In supplementary, we discuss that shifting the parameters of various generator layers results in different effects, i.e., earlier layers induce global geometrical transformations. The middle layers correspond to localized transformations of object parts, and the final layers are responsible for coloring manipulations.

\subsection{Comparison}
\label{sect:comparison}

In this subsection, we compare different techniques of parameter space navigation, both qualitatively and quantitatively. Specifically, we compare the following methods:

\begin{itemize}
    \item \textbf{SVD} performs a smooth changing of a particular singular value in the SVD of a flattened generator layer. We include this method as a baseline inspired by the evidence from \cite{robb2020few}, which shows that varying the singular values often result in interpretable manipulations.
    
    \item \textbf{Optimization-based} method navigates $\Theta$ along the directions discovered as described in \sect{optimization}.
    
    \item \textbf{Spectrum-based} method navigates $\Theta$ along the directions discovered as described in \sect{spectrum}.
    
    \item \textbf{Hybrid} method navigates $\Theta$ along the directions discovered as described in \sect{hybrid}.
\end{itemize}

\begin{figure}
    \centering
    \includegraphics[width=\columnwidth]{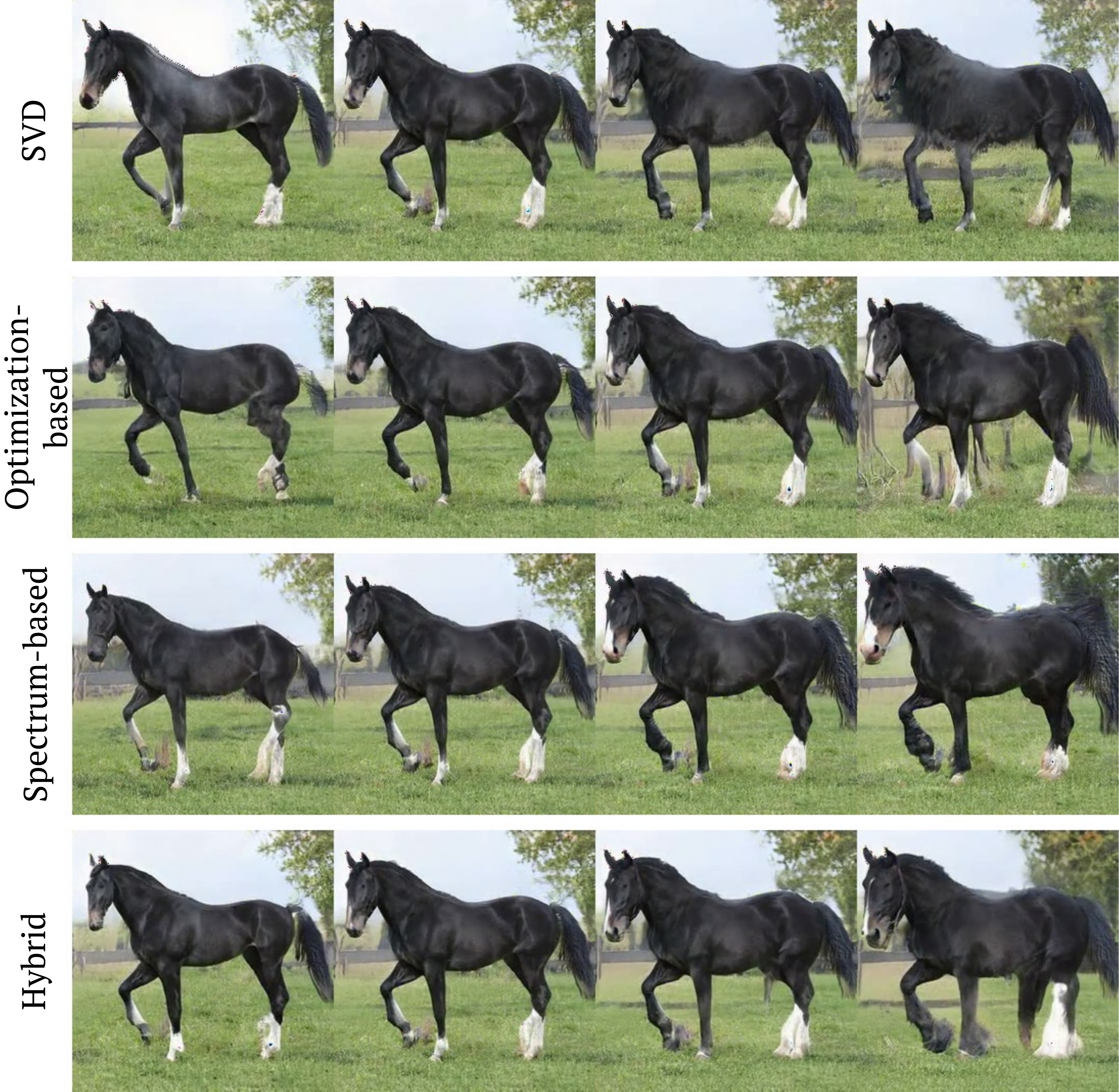}
    \caption{Comparison of the ``Thickness'' direction discovered by different methods applied to the 3-rd conv layer of StyleGAN2.}
    \label{fig:horse_cmp}
    \vspace{-2mm}
\end{figure}

\begin{figure}
    \centering
    \includegraphics[width=\columnwidth]{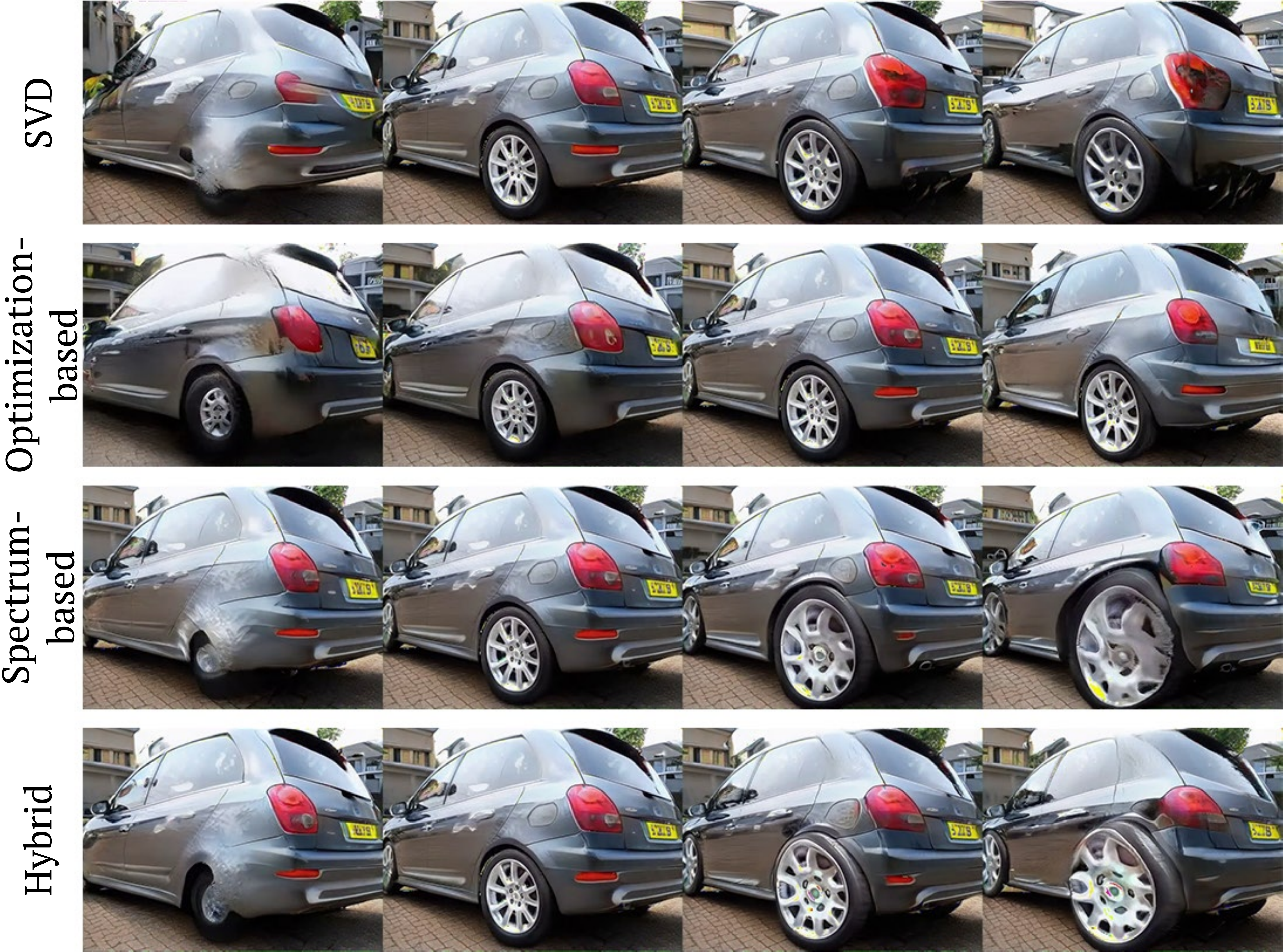}
    \caption{Comparison of the ``Wheel size'' direction discovered by different methods applied to the 5-th conv layer of StyleGAN2.}
    \label{fig:car_cmp}
    \vspace{-5mm}
\end{figure}

To compare the methods, we apply all of them to the particular layer. Interestingly, in some cases, all methods discover directions, roughly corresponding to the same visual effect. Given such a direction, we can qualitatively compare the methods by observing how this direction modifies the images produced from the same latent noise $z$. The typical visualizations are presented on \fig{horse_cmp} and \fig{car_cmp}. These examples show that the \textbf{SVD} baseline dramatically corrupts the image realism. In contrast, the techniques described in \sect{method} produce more clear and visible effects.

\begin{figure}
    \centering
    \includegraphics[width=\columnwidth]{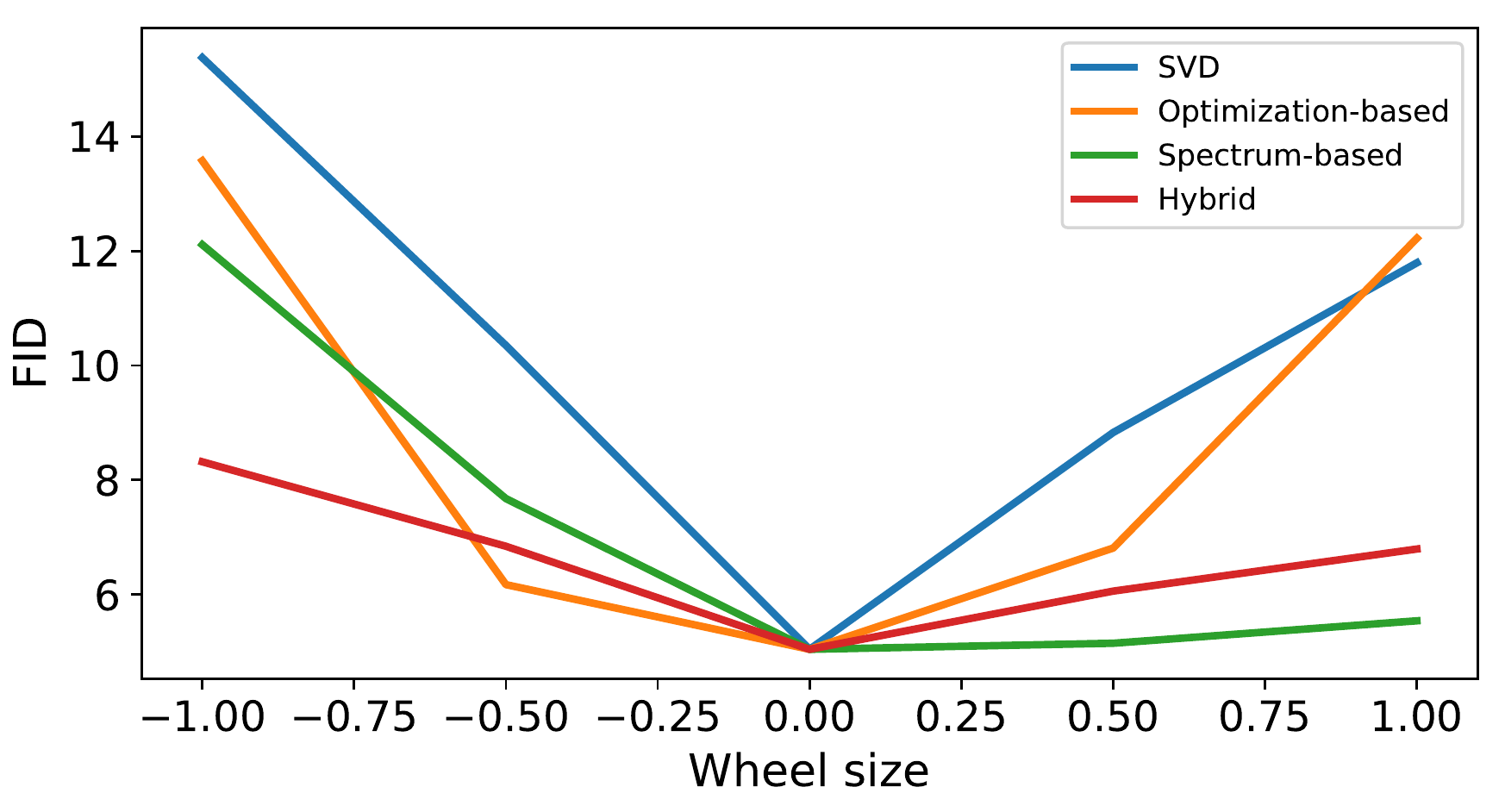}
    \caption{FID plots for the ``Wheel size'' visual effect produced by different methods.}
    \label{fig:fid_cmp}
    \vspace{-5mm}
\end{figure}

For a quantitative comparison, we perform the following experiment. For each of four methods, we consider a direction that corresponds to the  ``Wheel size'' visual effect. This effect is chosen since it was identified by all the methods. Then, for the generators $G_{\theta+t\xi}(z)$, we plot the FID curves by varying the shift magnitudes $t$\footnote{Since different methods have to use different scales of $t$ values to achieve the same degree of the visual effect, we first calibrate these scales manually to guarantee that the max values of $t$ for all methods correspond to the same wheel size.}. For each $t$ the FID is always computed with $5 \cdot 10^4$ real and generated images.
The plots are presented on \fig{fid_cmp}. The SVD baseline is inferior to three advanced methods from \sect{method}, while there is no clear winner among them. Overall, in practice we recommend to use all three proposed methods to explore the parameter space of a particular GAN since we observed that each method can reveal effects, missed by the others.

\subsection{Latent transformations cannot produce these visual effects}

Here, we show that the manipulations induced by parameter shifts cannot be achieved by changing the latent codes. 
Specifically, we perform the following experiment. Given a shift $\xi$ modifying a generator $G_{\theta} \to G_{\theta + \xi}$, we look for a latent shift $h$ such that $G_{\theta}(z + h) = G_{\theta + \xi}(z)$. To this end, we solve the optimization problem
\vspace{-1mm}
\begin{equation}
    \underset{h}{\text{min}} \; \mathbb{E}_z||G_{\theta}(z + h) - G_{\theta + \xi}(z)||_2^2
\end{equation}
\vspace{-5mm}

\begin{figure}
    \centering
    \includegraphics[width=\columnwidth]{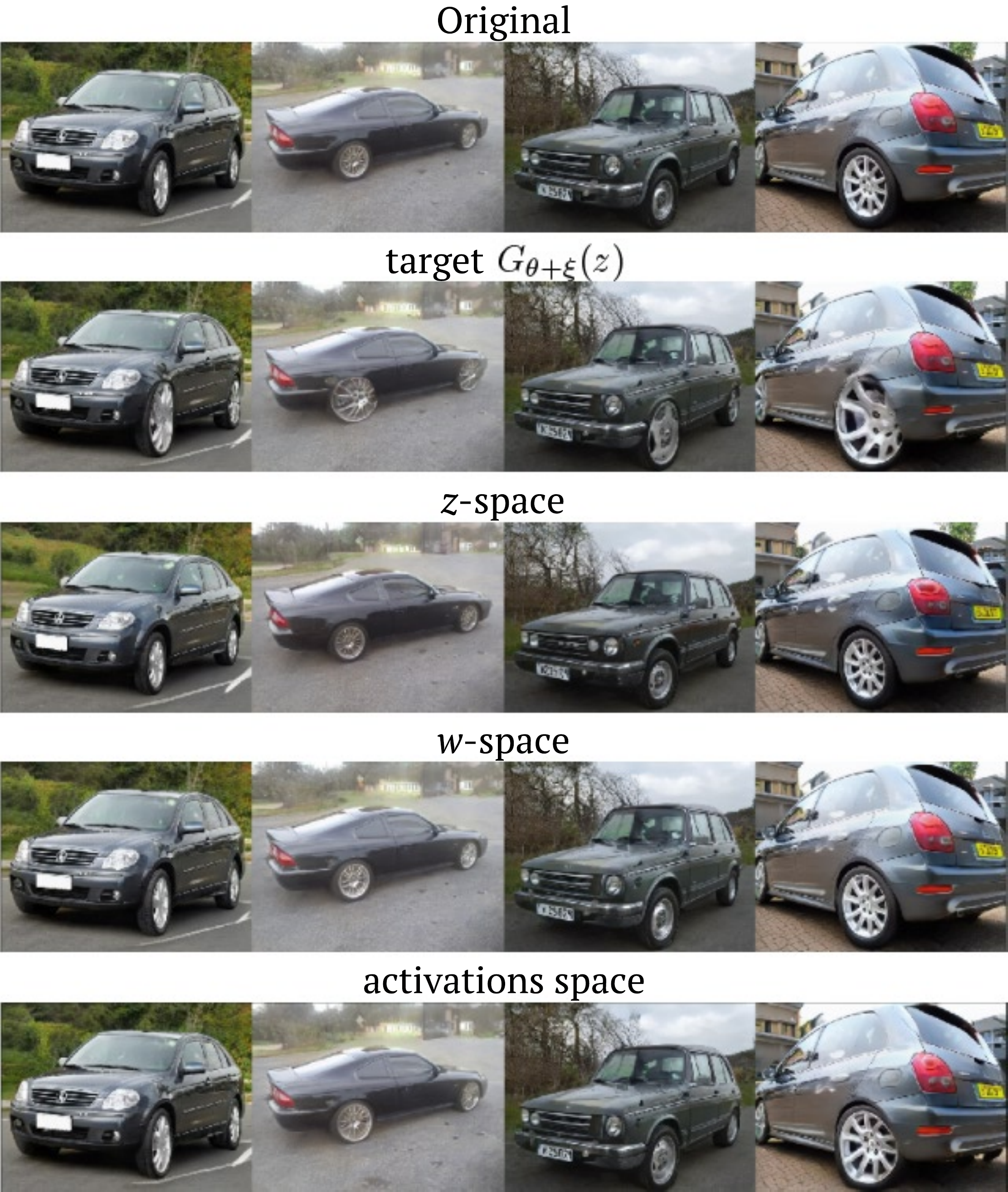}
    \caption{Unsatisfactory reproduction of the ``Wheel size'' manipulation by the shifts in the latent spaces and in the space of the intermediate activations.}
    \label{fig:steerability}
    \vspace{-2mm}
\end{figure}

For $G$ we take the StyleGAN2 pretrained on the LSUN-Car dataset and $\xi$ is the ``Wheels size'' direction. We optimize over $h$ from the input $\mathcal{Z}$-space, the $\mathcal{W}$-space and the space of the activation tensors before the modified convolutional layer. In all three settings, the optimization failed to obtain a latent shift that would properly mimic the visual effect. The results of the optimization are shown on \fig{steerability}. Similar results were obtained for other generators and directions reported in the paper, see supplementary. Thus, our approach reveals new semantic manipulations unattainable by existing methods.

\subsection{Editing real images}

\begin{figure}
    \centering
    \includegraphics[width=\columnwidth]{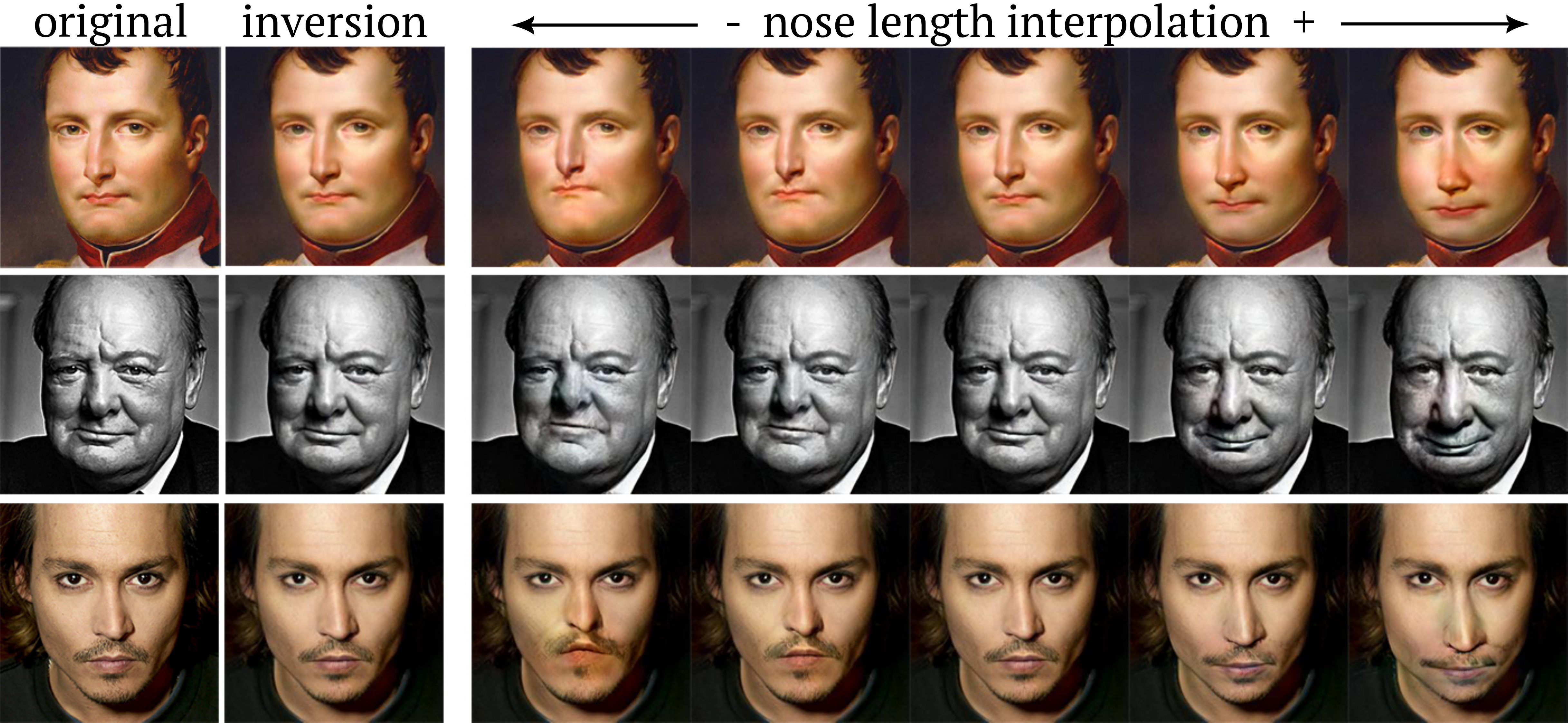}
    \caption{Interpolations along the ``Nose length'' direction for real images embedded in $\mathcal{W+}$ space.}
    \label{fig:real_nose}
    \vspace{-5mm}
\end{figure}

\begin{figure}
    \centering
    \includegraphics[width=\columnwidth]{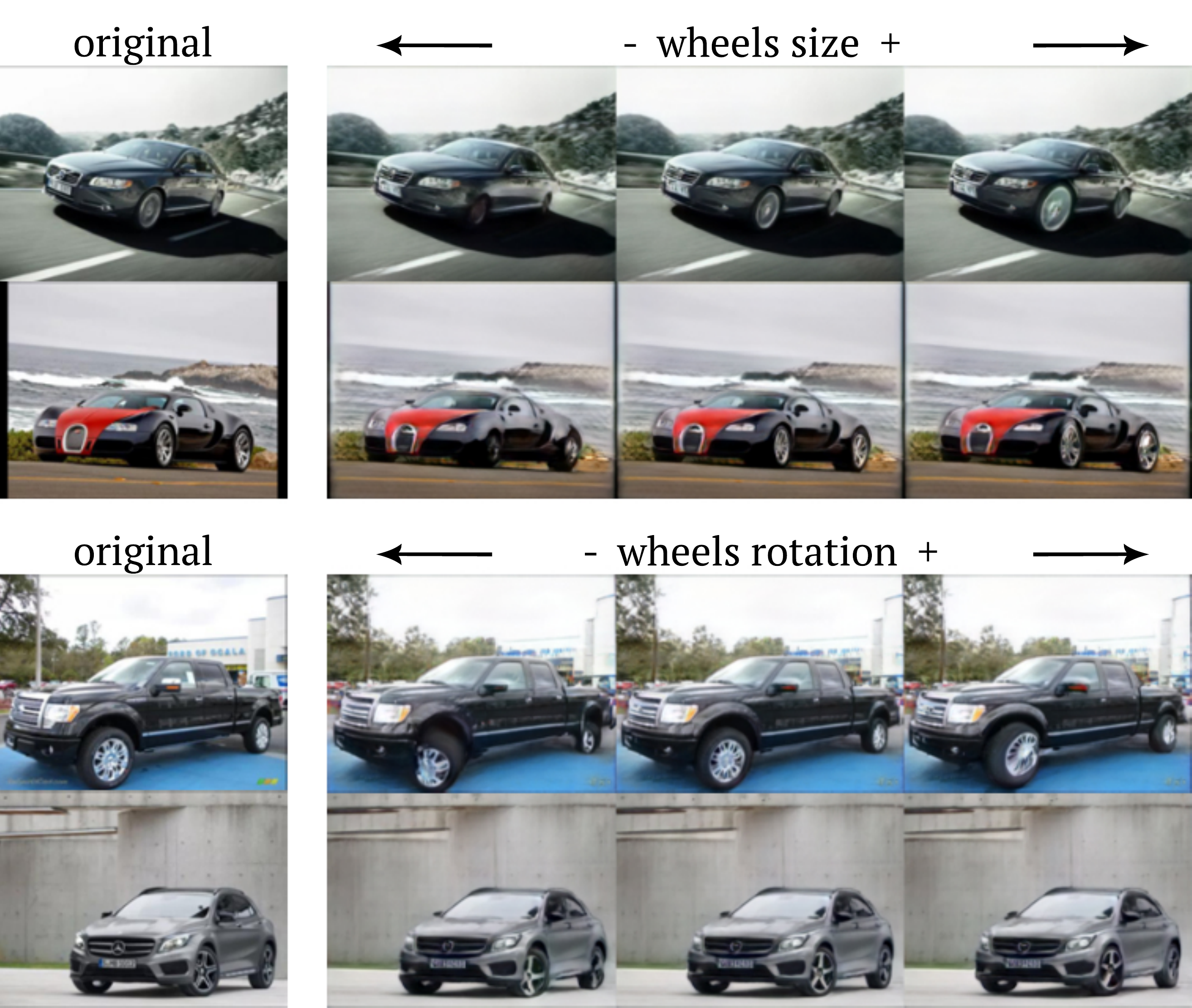}
    \caption{Manipulations of real car images embedded in $\mathcal{W+}$~space.}
    \label{fig:car_real}
    \vspace{-2mm}
\end{figure}

Importantly, the discovered image manipulations can be naturally applied to real images using the GAN inversion techniques, which embed a given image to the latent space of a pretrained GAN. Furthermore, the manipulations can be applied to the images embedded into the extended $\mathcal{W}+$ space of StyleGAN2. We argue that this is an appealing feature of our transformations since the conventional latent shifts require embeddings to the lower-dimensional $\mathcal{W}$-space, which can harm the inversion quality. To illustrate how the discovered visual effects perform on real images domain, we invert the real images with the standard StyleGAN2 projector in the $\mathcal{W}+$ space \cite{Karras2019stylegan2}. On the \fig{real_nose} we manipulate real images via the ``Nose length'' direction. \fig{car_real} illustrates different manipulations applied to real car images.

\subsection{Maintaining realism}

\begin{figure}
    \centering
    \includegraphics[width=\columnwidth]{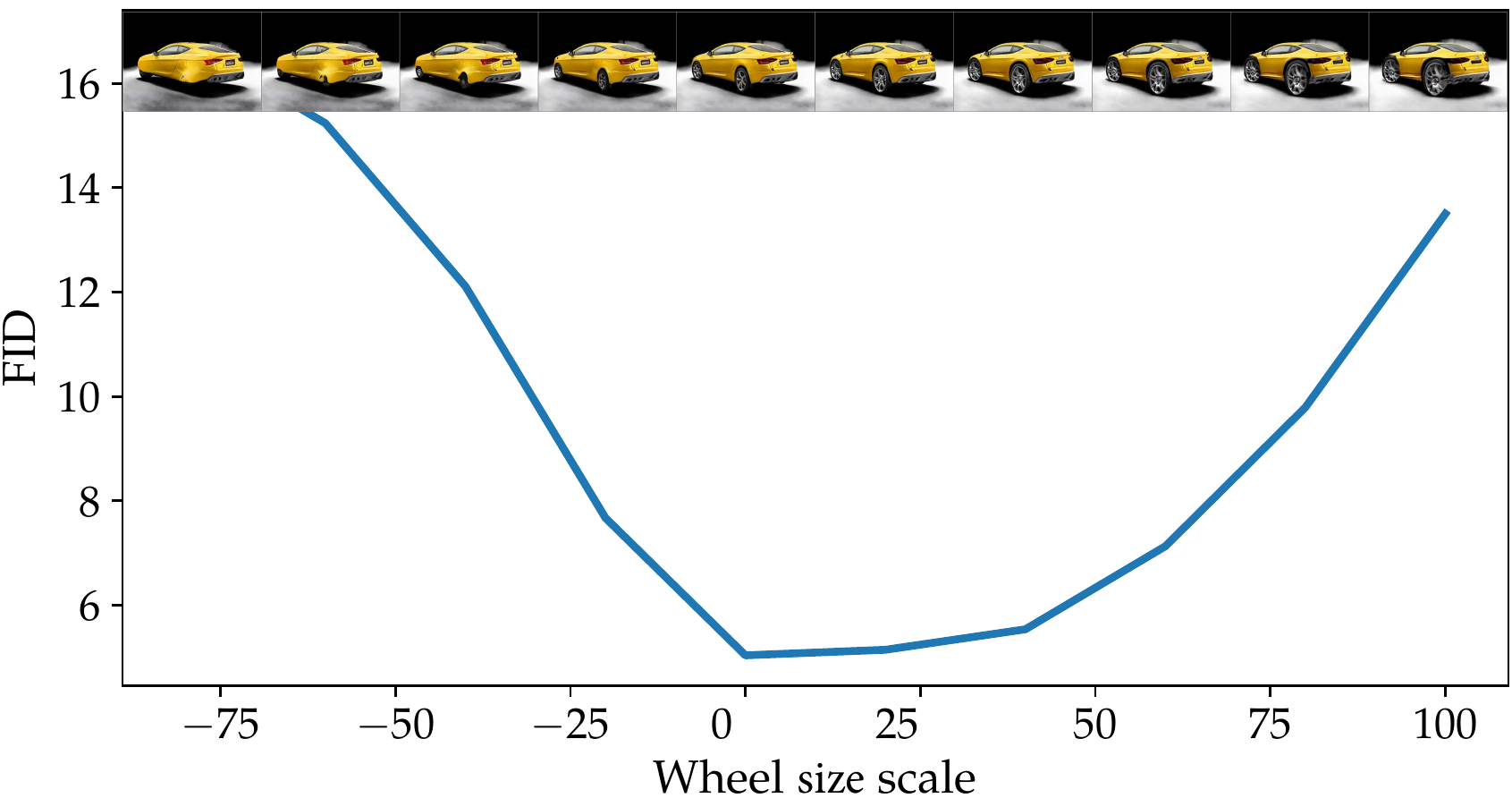}
    \includegraphics[width=\columnwidth]{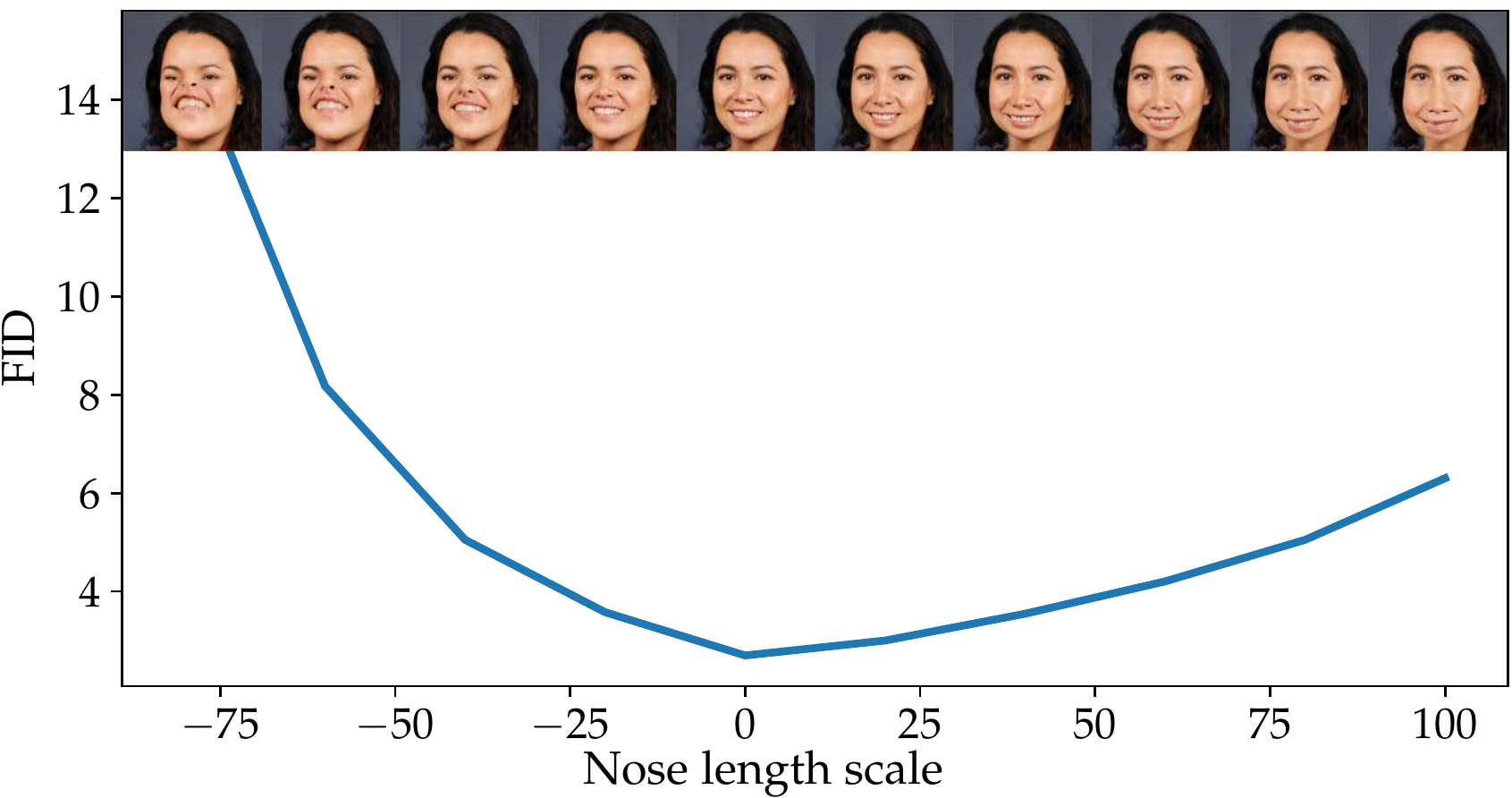}
    \caption{FID for the generators shifted along the ``Wheel size'' direction (\textit{top}) and the ``Nose length'' direction (\textit{bottom}).}
    \label{fig:fids_wheel}
    \vspace{-5mm}
\end{figure}

To verify that modifying the generator parameters does not significantly harm the visual quality, we compute the Fr\`echet Inception Distance (FID) \cite{heusel2017gans} values for the modified generators. Namely, we plot the FID for the generators with parameters shifted with different magnitudes along the ``Nose length'' and ``Wheel size'' directions, see \fig{fids_wheel}.

Notably, even under extreme shift magnitudes, the manipulated samples have high visual quality. We also observe that the FID plot is not symmetric. We attribute this behavior to the fact that the real data can be inherently asymmetric with respect to specific attributes. For instance, the generated cars FID grows much faster once the wheels become small rather than large, probably, because LSUN-Cars contains cars with big wheels, but there are no cars without wheels. The same behavior appears with a human nose as tiny noses are rarer than long noses. We also present similar plots for other directions and datasets in the supplementary.

\subsection{Locality of visual effects}

\begin{figure}
    \centering
    \includegraphics[width=\columnwidth]{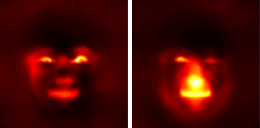}
    \caption{Averaged heatmaps of the pixel differences between the original and the edited images for the ``Eyes distance'' direction (\textit{left}) and the ``Nose length'' direction (\textit{right}).}
    \label{fig:heatmap}
\end{figure}

To illustrate that the navigation along the discovered directions results in an isolated and ``disentangled'' effect, we perform the following. For a particular direction $\xi$, we compute the per-pixel differences $\|G_{\theta + t \cdot \xi}(z) - G_{\theta}(z)\|_2^2$ averaged over $1600$ $z$-samples and $20$ shift magnitudes $t$ from a uniform grid in a range $[-100, 100]$. \fig{heatmap} shows the averaged heatmap for the ``Eyes distance'' and ``Nose length'' directions. Notably, the ``Nose length'' slightly affects the eyes and mouth since the extreme shift pushes the nose to overlap these face areas.

\begin{figure}
    \centering
    \includegraphics[width=\columnwidth]{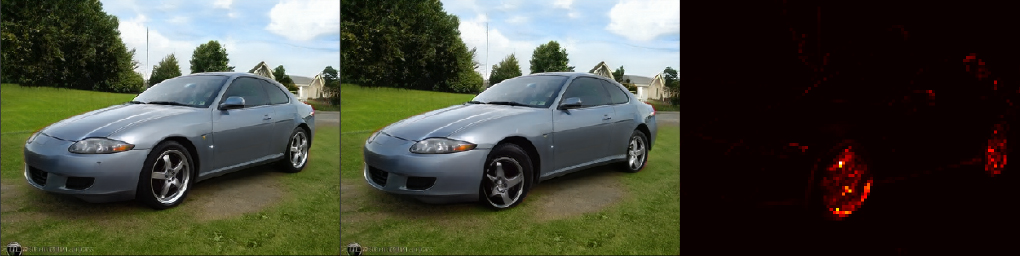}
    \caption{\textit{Left}: original image; \textit{center}: a shift in the direction ``Wheel rotation''; right: the squared distances between the original and edited image averaged over different shift magnitudes.}
    \label{fig:wheel_heatmap}
    \vspace{-3mm}
\end{figure}

As another example, \fig{wheel_heatmap} shows the averaged square distance between an image generated by the original StyleGAN2 and its shifts along the ``wheel rotation'' direction. Only the ``wheel regions'' are affected.

\subsection{Alternative GANs models}
The proposed method is model-agnostic and can be applied even to image-to-image models. In the supplementary we present qualitative results of our approach been applied to pix2pixHD \cite{wang2018pix2pixHD} pretrained on Cityscapes dataset and BigGAN \cite{brock2018large} pretrained on Imagenet.

\section{Conclusion}
\label{sect:conclusion}

In this paper, we have investigated the possibilities for GAN-based semantic editing via smooth navigation in the space of the generator's parameters. In particular, we have shown that this space contains various interpretable controls, which can be identified via efficient and straightforward procedures. Given the simplicity and universality of the proposed techniques, these controls become a valuable complement to existing visual editing tools. 

{\small
\bibliographystyle{ieee_fullname}
\bibliography{egbib}
}

\onecolumn
\twocolumn
\section{Supplementary Material}
\subsection{Discovered visual effects}

\begin{figure}
    \vspace{-5mm}
    \centering
    \includegraphics[width=\tablescale\textwidth]{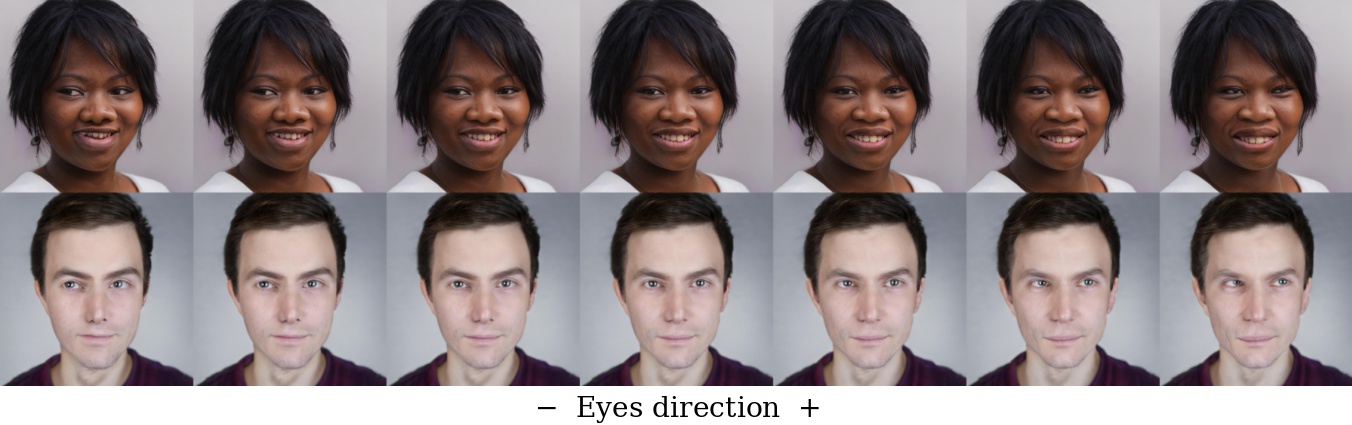}
    \includegraphics[width=\tablescale\textwidth]{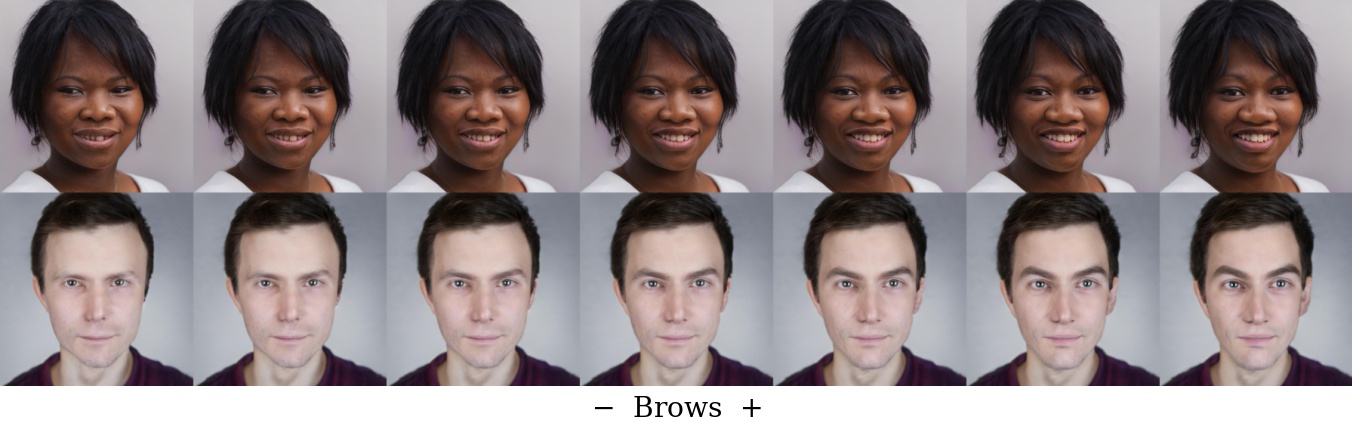}
    \includegraphics[width=\tablescale\textwidth]{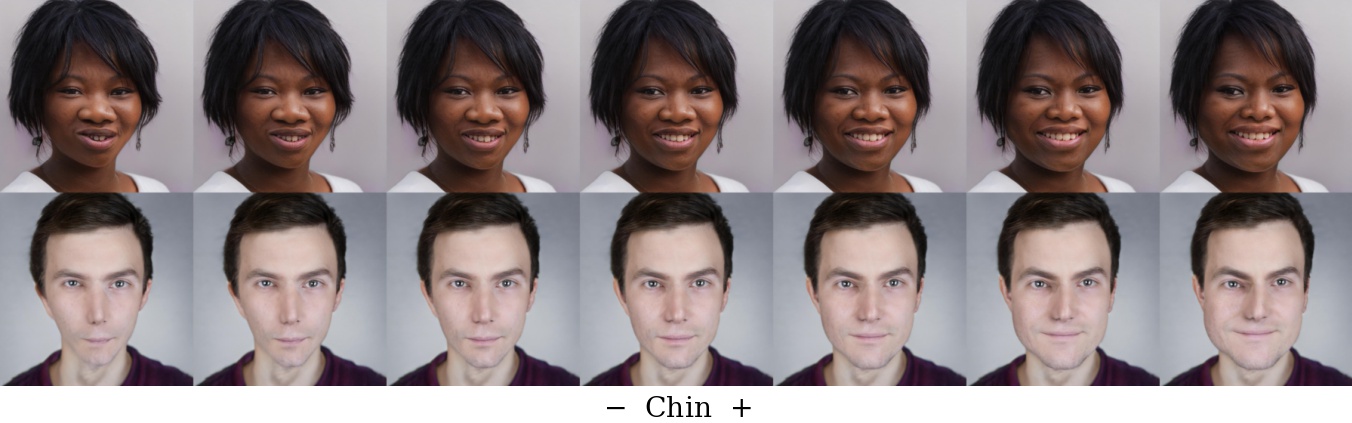}
    \includegraphics[width=\tablescale\textwidth]{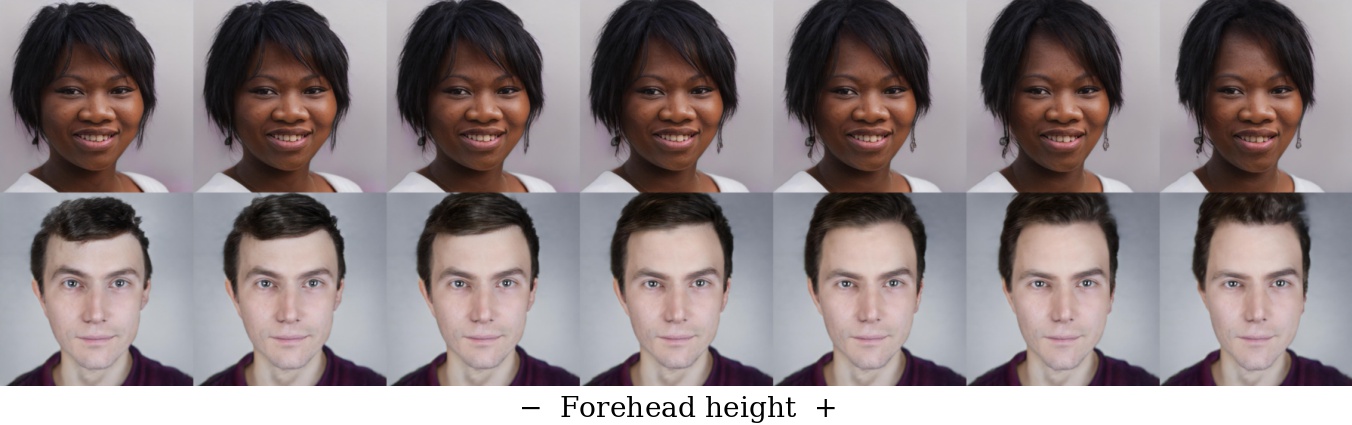}
    \includegraphics[width=\tablescale\textwidth]{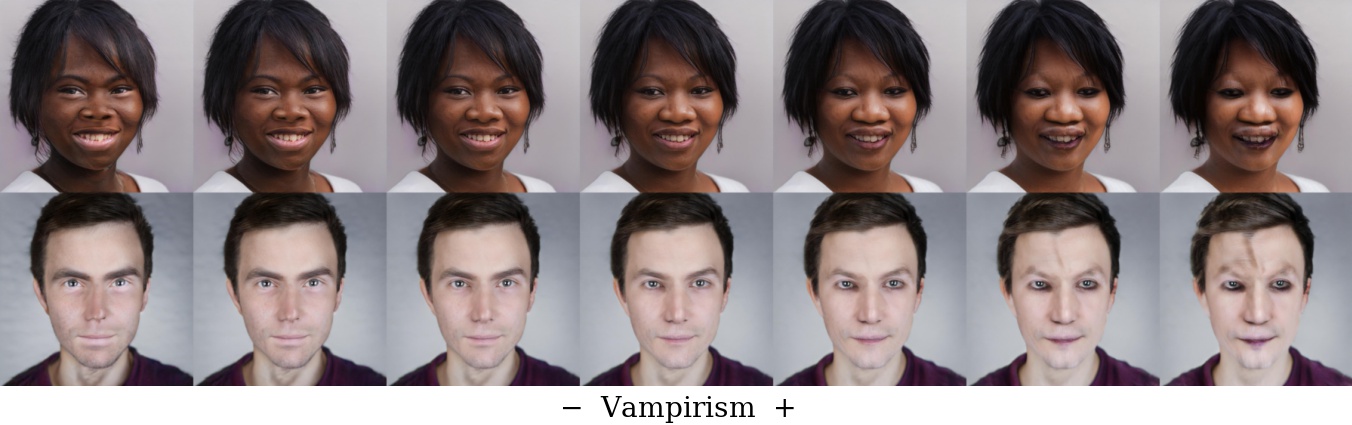}
    \caption{Examples of effects discovered for FFHQ StyleGAN2.}
    \label{fig:ffhq}
\end{figure}

\begin{figure}
    \vspace{-20mm}
    \centering
    \caption{Examples of effects discovered for LSUN-Church StyleGAN2.}
    \includegraphics[width=\tablescale\textwidth]{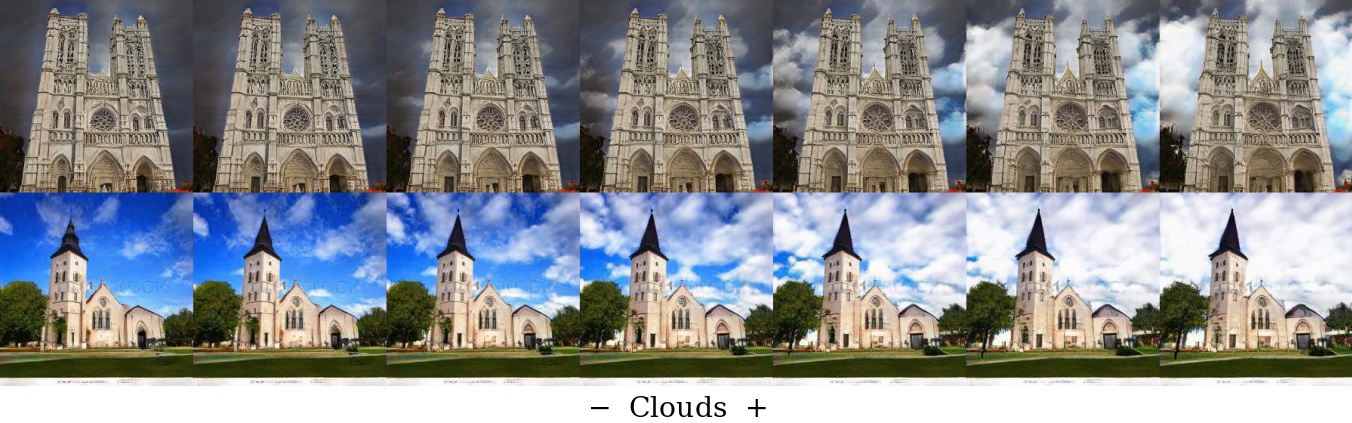}
    \includegraphics[width=\tablescale\textwidth]{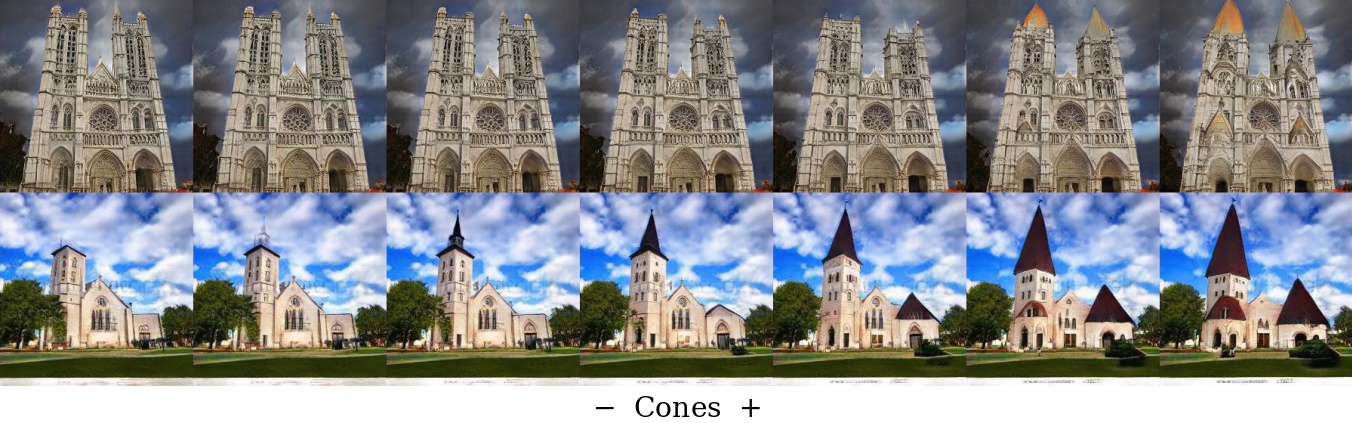}
    \includegraphics[width=\tablescale\textwidth]{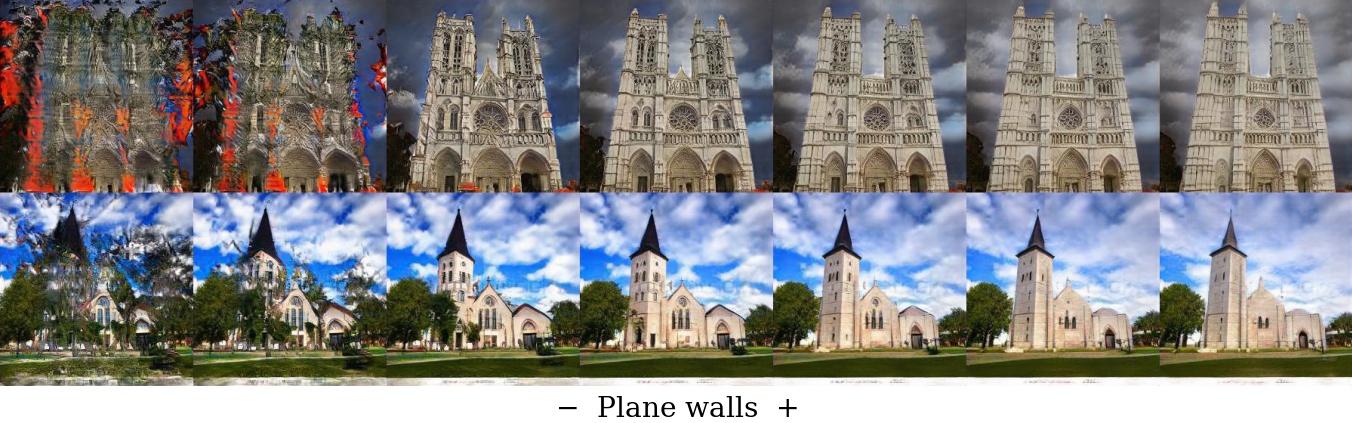}
    \label{fig:church}

    \centering
    \caption{Examples of effects discovered for LSUN-Car StyleGAN2.}
    \includegraphics[width=\tablescale\textwidth]{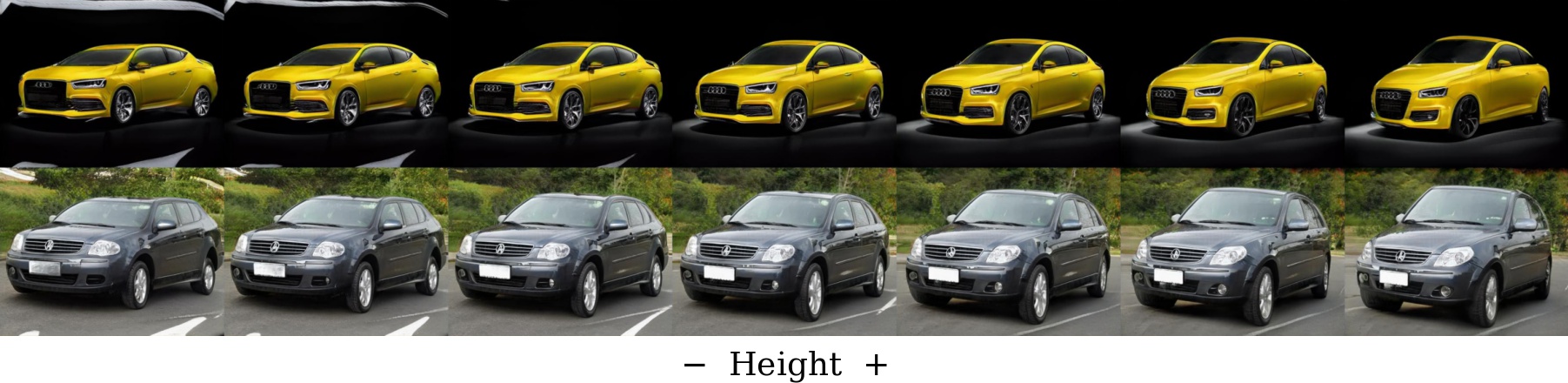}
    \includegraphics[width=\tablescale\textwidth]{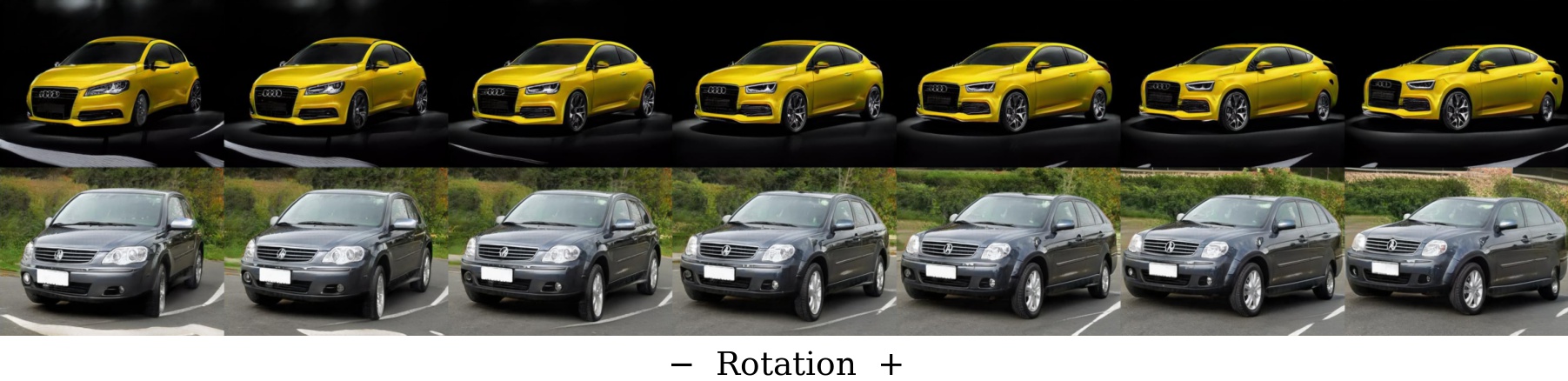}
    \includegraphics[width=\tablescale\textwidth]{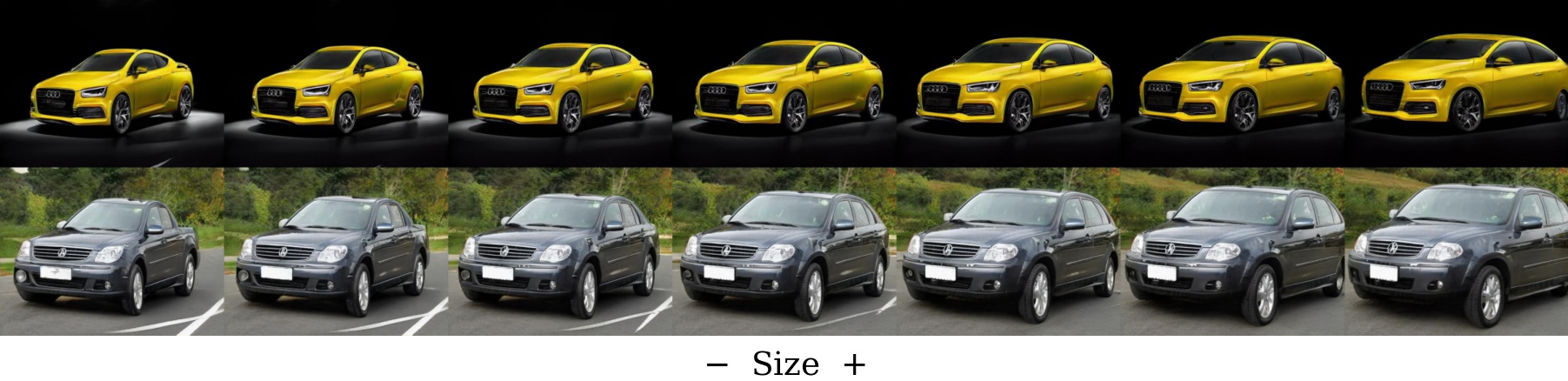}
    \label{fig:car}
\end{figure}

\fig{ffhq}, \fig{church}, \fig{car}, \fig{depth_ablation} present more examples of visual effects achievable by navigating the StyleGAN2 parameter space. Additional examples for four datasets are provided in the GitHub repository\footnote{\url{https://github.com/yandex-research/navigan}} along with the PyTorch implementation of our method.

\subsection{Dependence on the layer depth}

\begin{figure}
    \vspace{-15mm}
    \centering
    \includegraphics[width=\tablescale\textwidth]{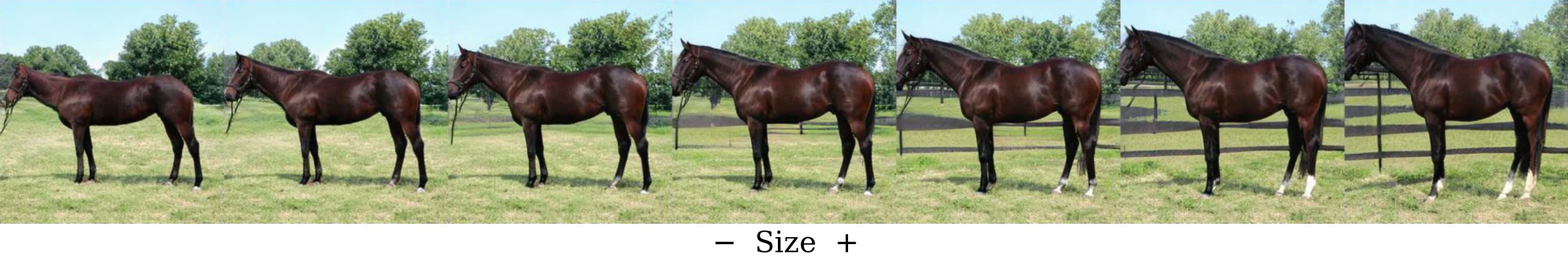}
    \includegraphics[width=\tablescale\textwidth]{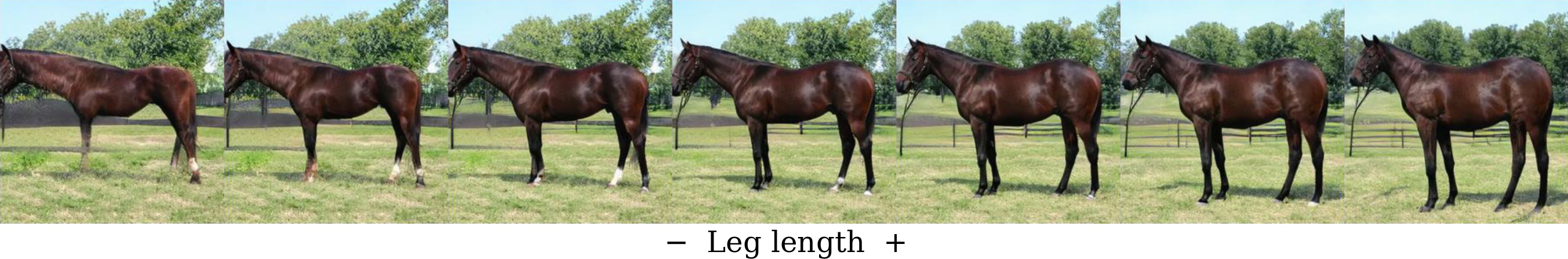}
    \includegraphics[width=\tablescale\textwidth]{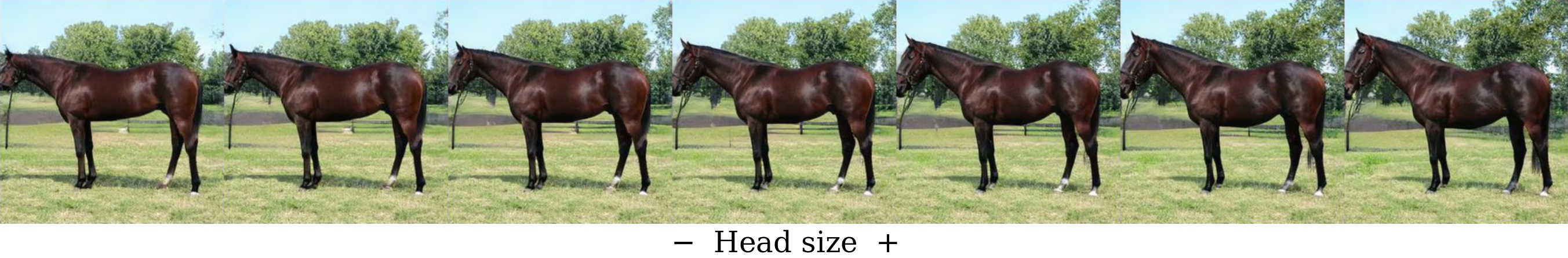}
    \includegraphics[width=\tablescale\textwidth]{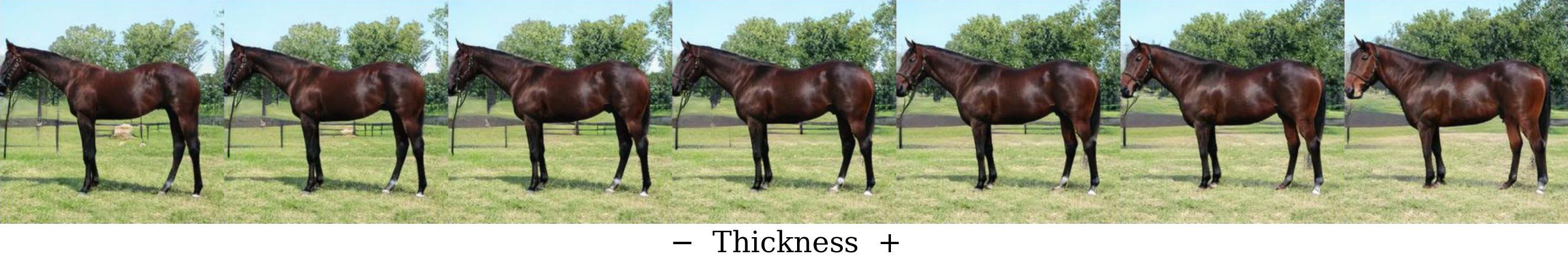}
    \includegraphics[width=\tablescale\textwidth]{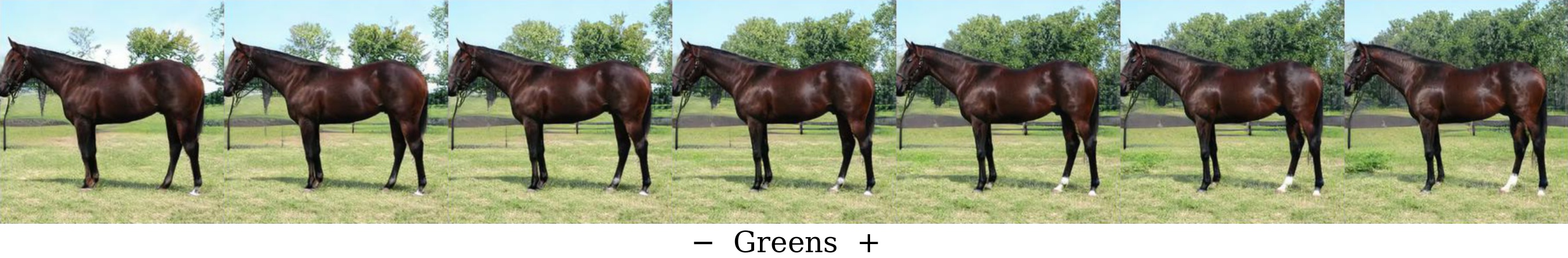}
    \includegraphics[width=\tablescale\textwidth]{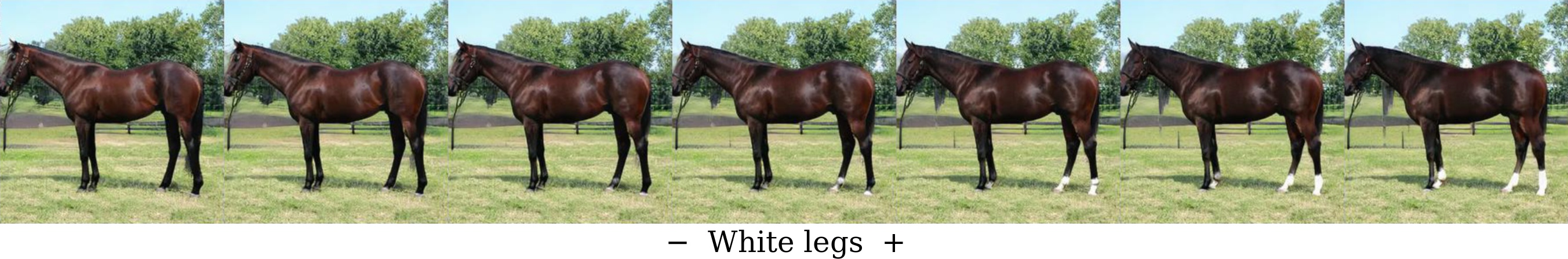}
    \includegraphics[width=\tablescale\textwidth]{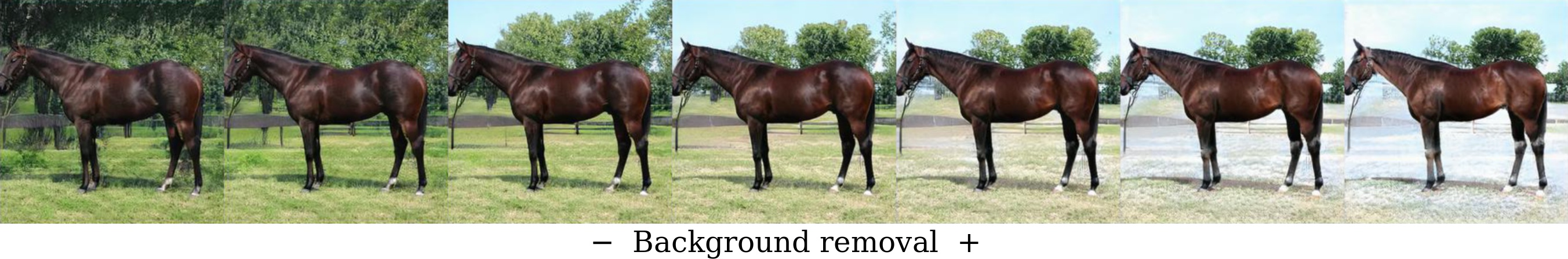}
    \includegraphics[width=\tablescale\textwidth]{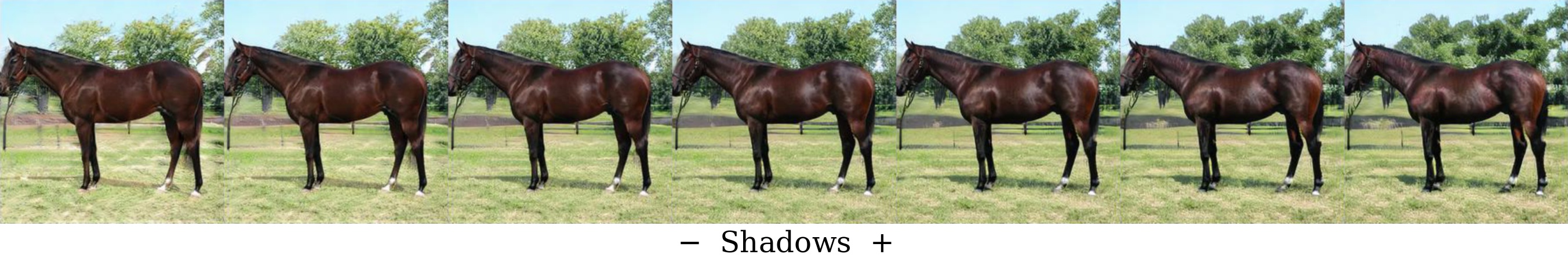}
    \includegraphics[width=\tablescale\textwidth]{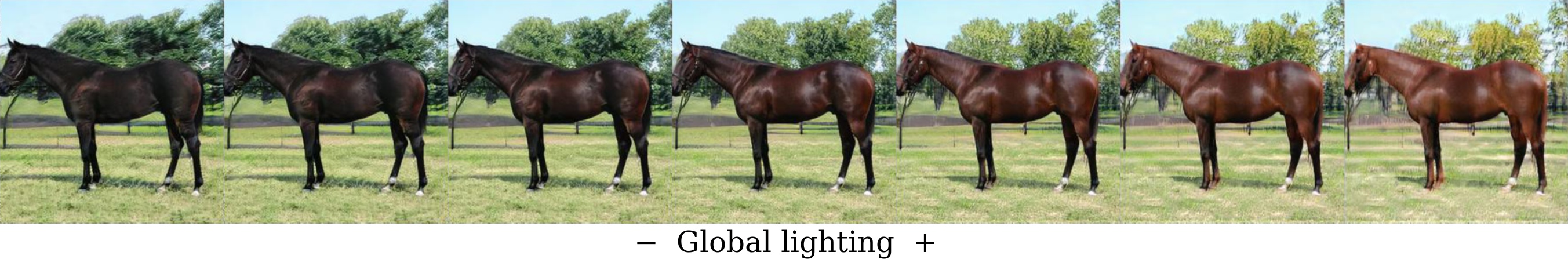}
    \includegraphics[width=\tablescale\textwidth]{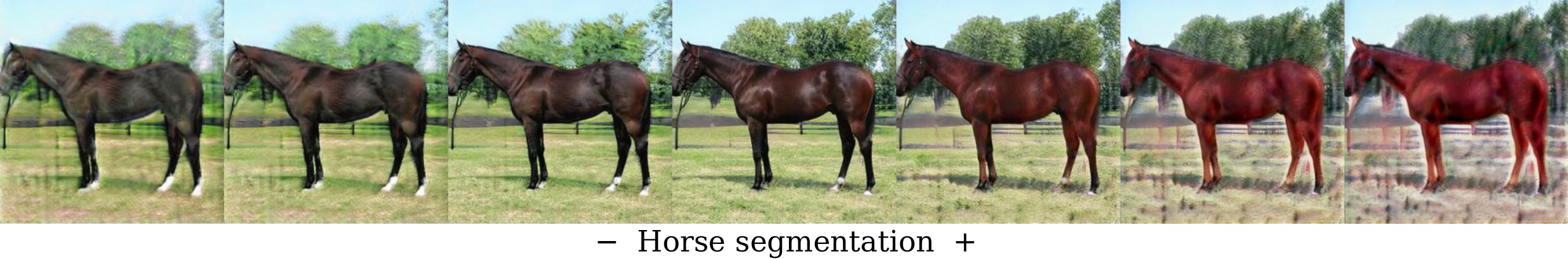}
    \caption{The effects discovered for the different layers of StyleGAN2 trained on the LSUN-Horse dataset. Each row corresponds to the particular StyleGAN2 generator layer.}
    \label{fig:depth_ablation}
\end{figure}

Different generator layers were shown to capture different image properties \cite{bau2018gan}. Accordingly, navigating the parameter space of layers from different depths also discovers the effects of different types. For the LSUN-Horse dataset, \fig{depth_ablation} visualizes the interpretable manipulations discovered at different depths, one manipulation per each StyleGAN2 layer. Notably, the earlier layers are generally responsible for global geometric transformations (size, leg length). Then, the intermediate layers typically result in more localized geometric manipulations (head size, thickness). They are followed by localized color manipulations (greens, white legs, background removal, shadows). The last layers correspond to global lighting effects (global lighting, horse reddening). Here we do not consider several final layers since they capture only trivial color-editing transformations. On other datasets, the distribution of typical effects over different layers is mostly the same.

\subsection{Comparison of approaches}

For a more quantitative comparison of the four approaches, we apply all of them to the fourth layer of the LSUN-Horse StyleGAN2 and manually annotate the controls discovered by each approach. For a fair comparison, each approach was set to discover $K{=}64$ directions. The result of the comparison is presented in \tab{horse_cmps}. If different approaches reveal directions with the same semantic meaning, we underline the best of them, which corresponds to the most clear and disentangled effect. Overall, the hybrid scheme performs best, both in terms of the number of discovered effects and their visual quality.

\begin{table}[h]
\centering
\begin{tabular}{ | c | m{4.5cm} | } 
\hline
SVD & thickness \\ 
\hline
Optimization-based & thickness, rotation, legs distance \\ 
\hline
Spectrum-based & thickness, rotation, head size, body-head proportion, vertical shift \\ 
\hline
Hybrid & \underline{thickness}, rotation, legs distance, body-head proportion, head rotation, belly size\\
\hline
\end{tabular}
\vspace{2mm}
\caption{Directions discovered by four methods by navigating the subspace of parameters for the fourth layer of LSUN-Horse StyleGAN2.}
\label{tab:horse_cmps}
\end{table}

\subsection{Further experiments}
\begin{figure}[h!]
    \includegraphics[width=\columnwidth]{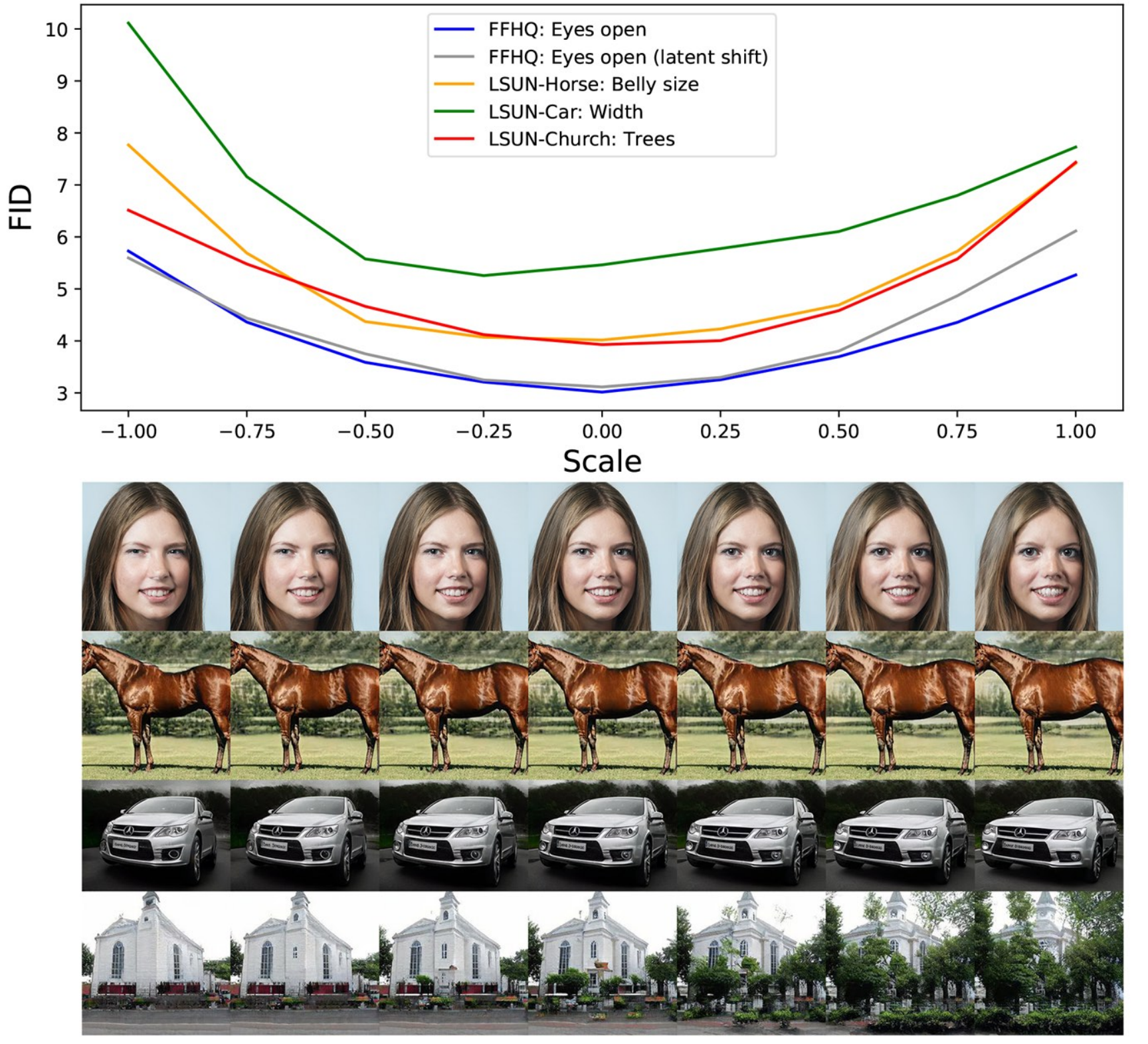}
    \caption{FID values for different weights shifts scales for some of the discovered directions. We also depict this plot for the ``Opened eyes'' direction discovered with the GANSpace in latent space.}
    \label{fig:interp_fids}
\end{figure}

\begin{figure}[h!]
    \centering
    \includegraphics[width=0.7\columnwidth]{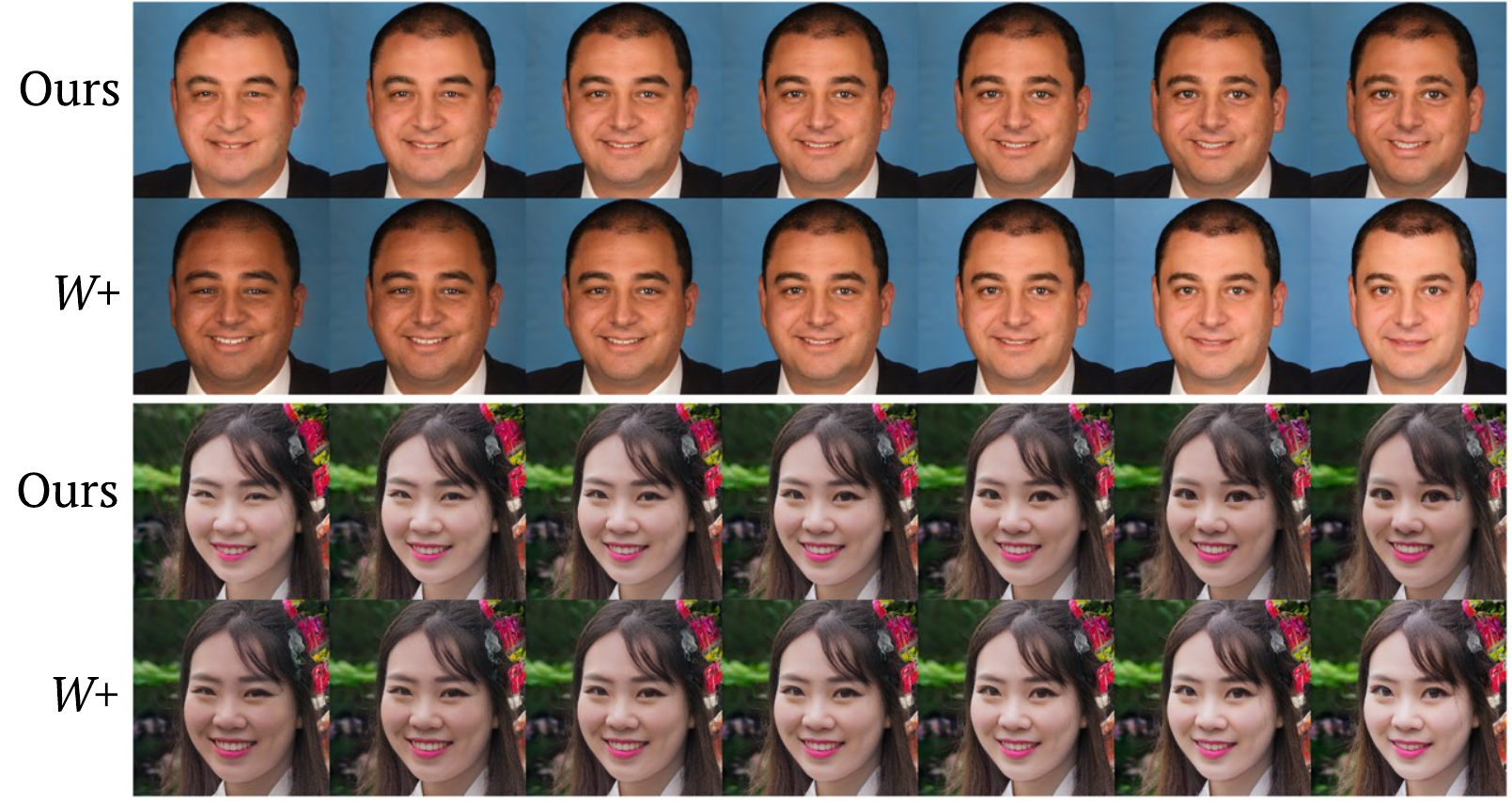}
    \caption{Comparition of the ``Opened eyes'' direction discovered with our approach and the GANSpace method in $\mathcal{W}+$ }
    \label{fig:ganspace_eyes_cmp}
\end{figure}

\begin{figure}[h!]
    \centering
    \captionsetup{width=0.7\linewidth}
    \includegraphics[width=0.49\columnwidth]{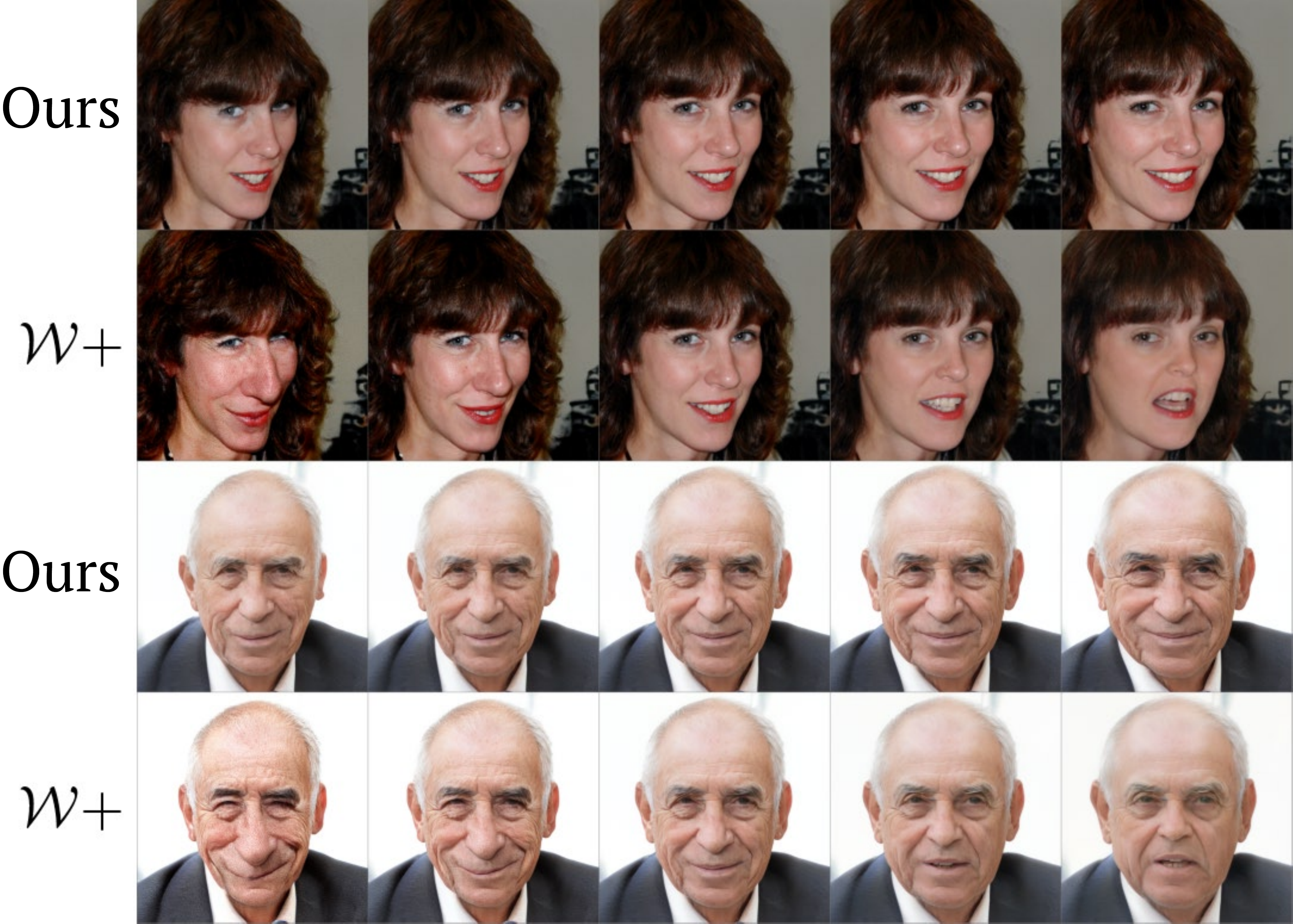}\vspace{2mm}
    \includegraphics[width=0.49\columnwidth]{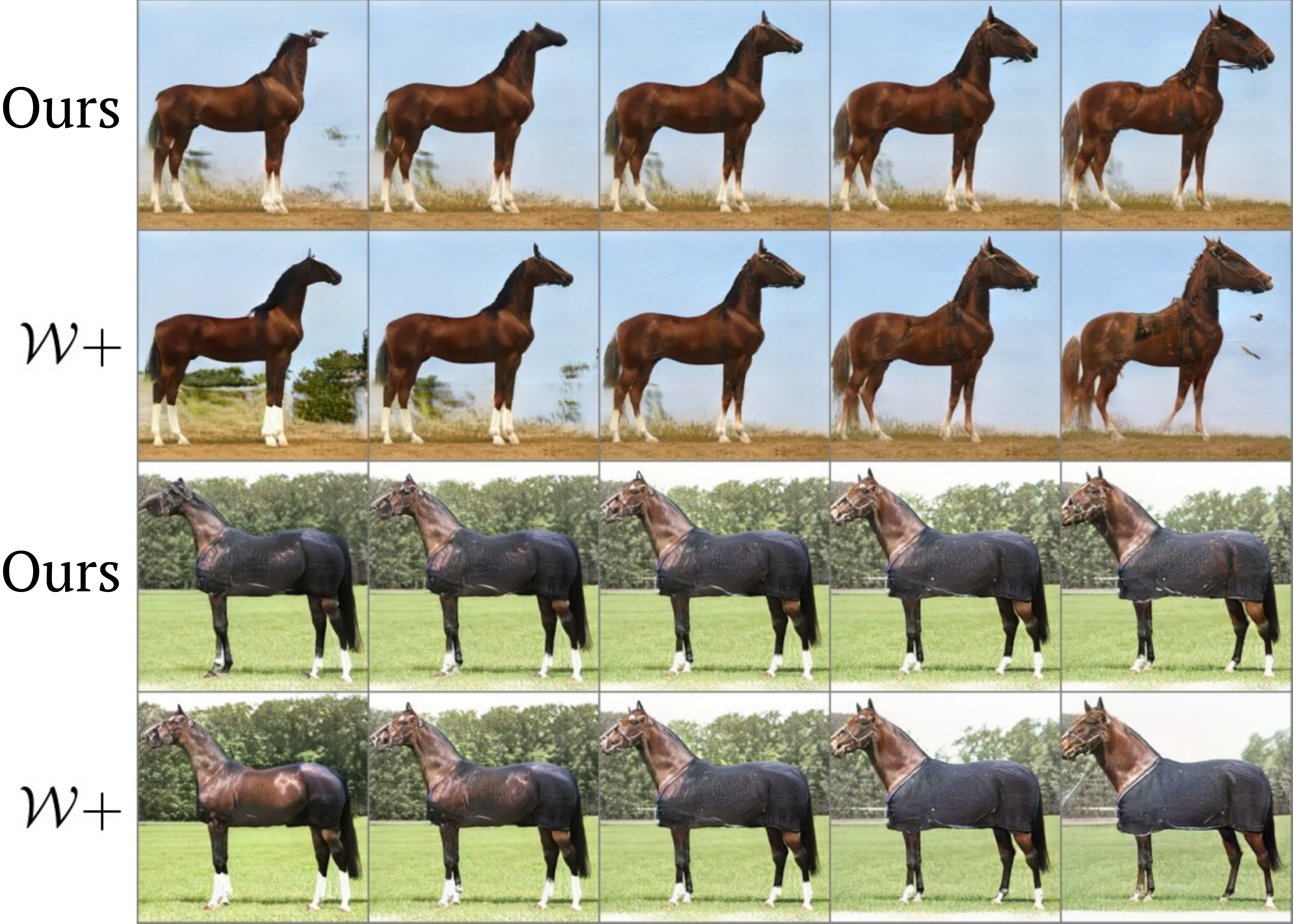}\vspace{2mm}
    \includegraphics[width=0.49\columnwidth]{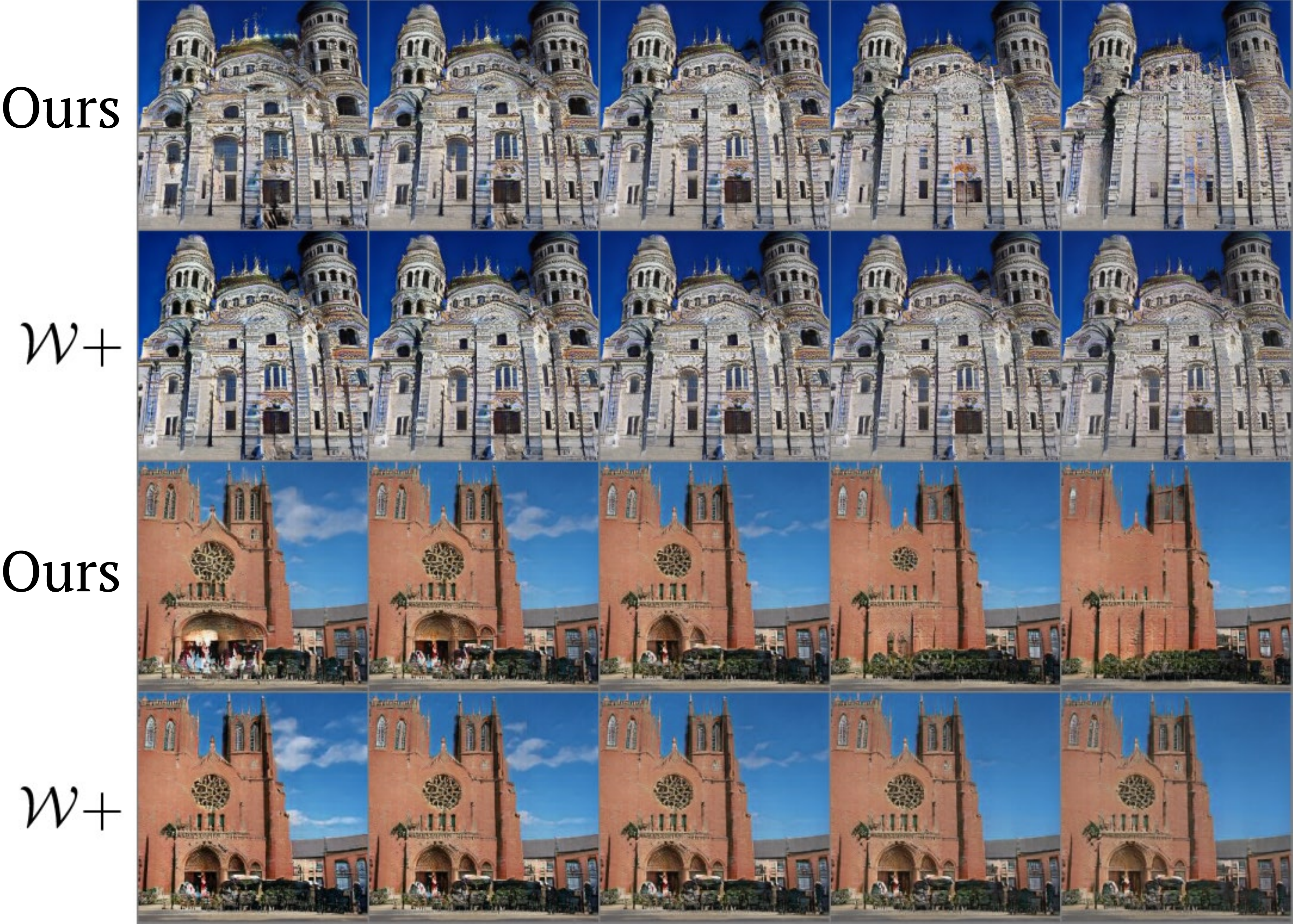}
    \caption{Unsatisfactory reproduction of ``Face width'', ``Horse head size'' and ``Plane walls'' manipulations by the shifts in $\mathcal{W}+$.}
    \label{fig:cmp_w_plus}
\end{figure}

\fig{interp_fids} demonstrates the FID values for different weights shift amplitudes. We also plot the FID values for one of the directions discovered with the GANSpace \cite{harkonen2020ganspace} approach in the latent space. The comparison of transformations induced by this latent shift and the weights shift is presented on \fig{ganspace_eyes_cmp}. \fig{cmp_w_plus} demonstrates transformations induced by the weights shifts that are not reachable by $\mathcal{W}+$ shifts. As described in Section 4.3, we optimize these latent shifts to reproduce the weights shifts.

\subsection{Alternative GANs models}
As the proposed approach is model-agnostic, here we present qualitative results for different generators. We always apply our technique to convolutional weights of a particular layer and use the same hyperparameters as for StyleGAN2. Here we present some of the discovered transformations for pix2pixHD \cite{wang2018pix2pixHD} pretrained on Cityscapes \cite{Cordts2016Cityscapes} and for BigGAN \cite{brock2018large} pretrained on Imagenet. During training for pix2pixHD, we use the same input segmentation masks for the original image $G_\theta(\mathrm{mask})$ and the shifted one $G_{\theta + t \cdot \xi_k}(\mathrm{mask})$. For BigGAN, we pass samples pairs $G_\theta(z, c)$ and $G_{\theta + t \cdot \xi_k}(z, c)$ with the same class label $c$ picked uniformly from $\{1,\dots,1000\}$. 

\onecolumn
\pagenumbering{gobble}

\onecolumn

\begin{figure}
    \vspace{-15mm}
    \centering
    \includegraphics[width=0.85\columnwidth]{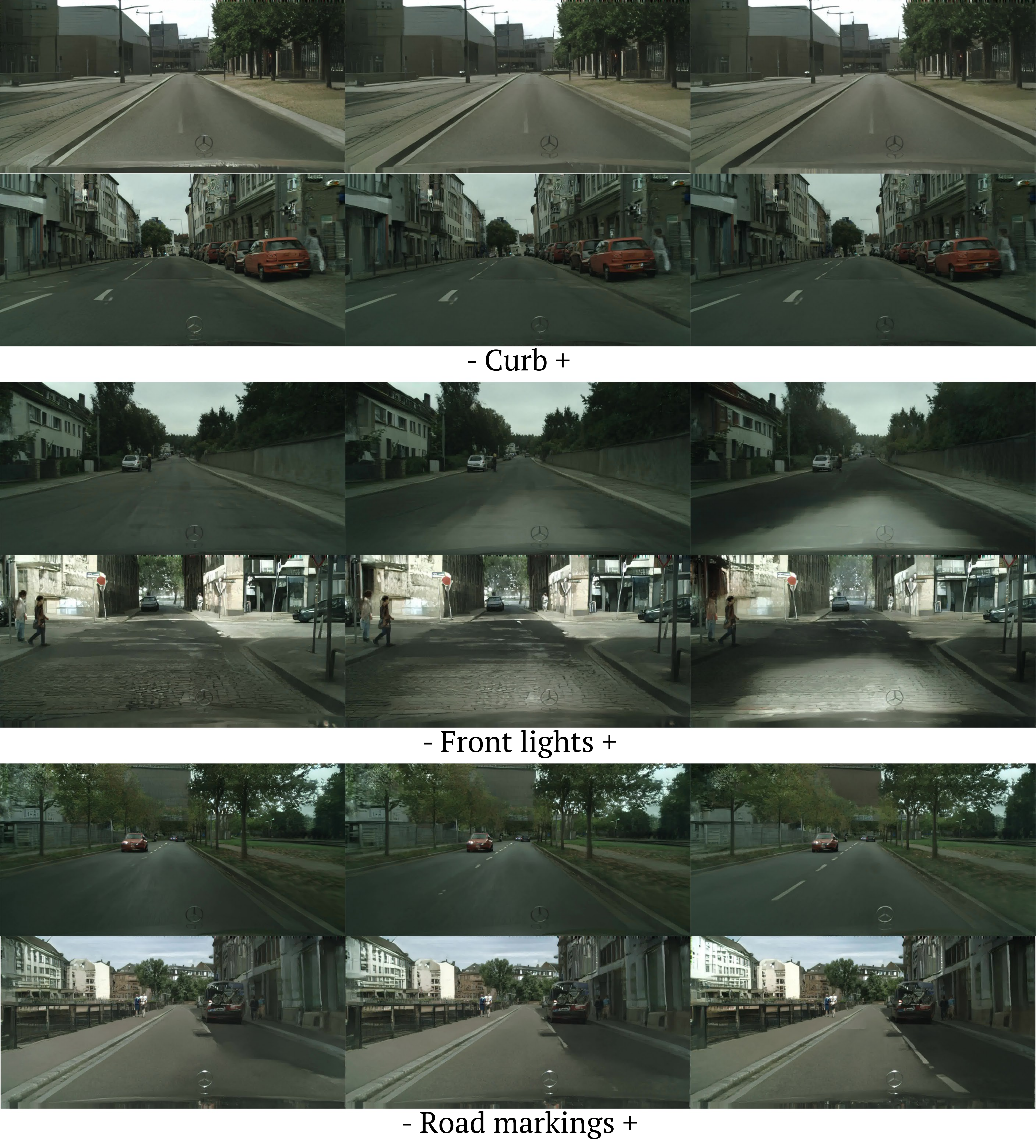}
    \includegraphics[width=0.85\columnwidth]{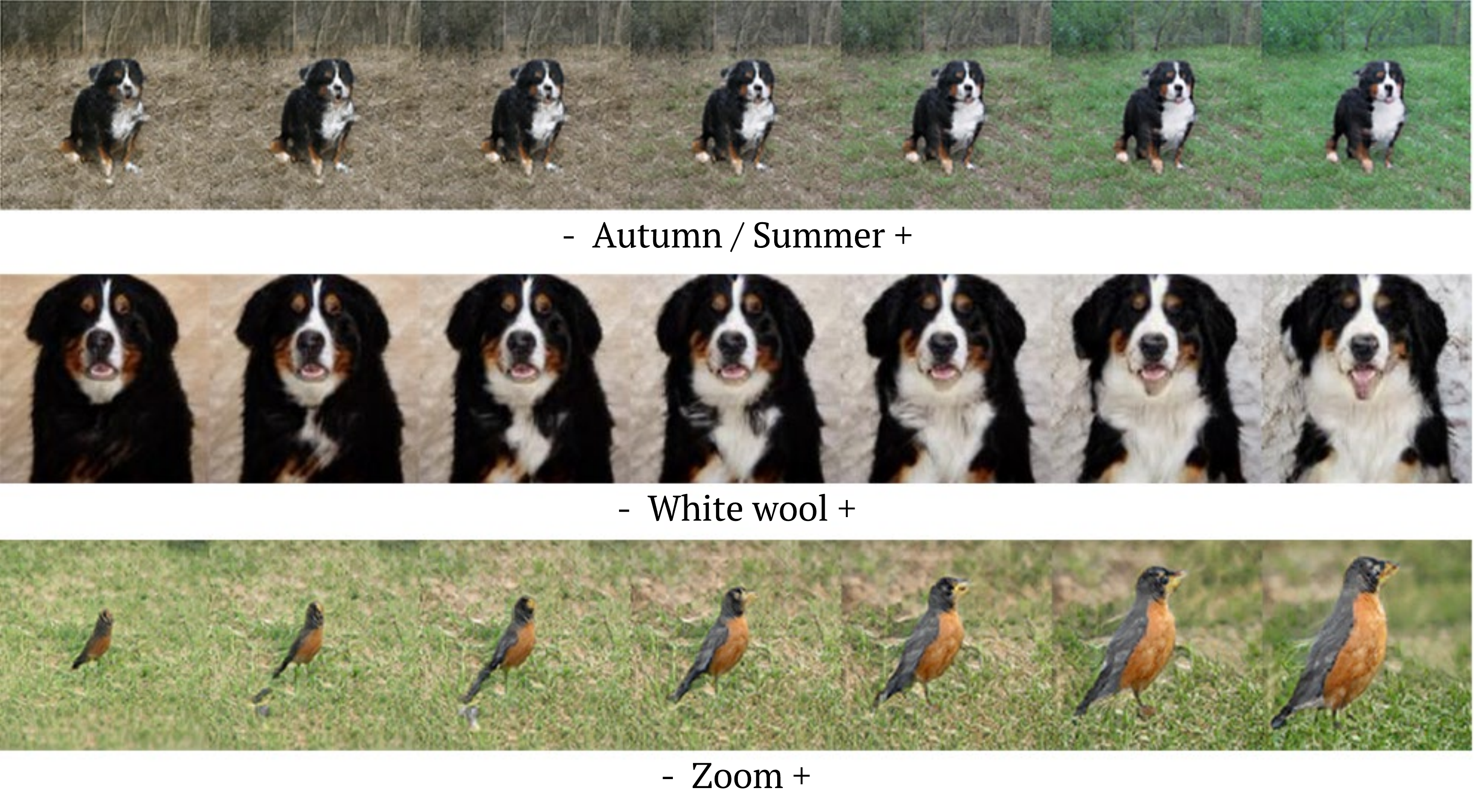}
    \caption{Some of the effects discovered for pix2pixHD model pretrained on Cityscapes (top 3) and BigGAN (bottom 3).}
    \label{fig:my_label}
\end{figure}

\end{document}